 \documentclass[pmlr,twocolumn,10pt]{jmlr} 

\usepackage{booktabs}
\usepackage{siunitx}

\usepackage[switch]{lineno}

\usepackage{subcaption}
\usepackage{color, colortbl}
\usepackage[font=footnotesize, skip=0.5pt]{caption}
\definecolor{LightGrey}{gray}{0.9}
\newcommand\numberthis{\addtocounter{equation}{1}\tag{\theequation}}

\theorembodyfont{\upshape}
\theoremheaderfont{\scshape}
\theorempostheader{:}
\theoremsep{\newline}

\jmlrvolume{297}
\jmlryear{2025}
\jmlrworkshop{Machine Learning for Health (ML4H) 2025} 

 \title[Federated Variational Inference for Bayesian Mixture Models]{Federated Variational Inference for Bayesian Mixture Models}

\author{\Name{Jackie Rao}
       \Email{jackie.rao@mrc-bsu.cam.ac.uk}\\ 
       \addr MRC Biostatistics Unit, University of Cambridge
      \AND
      \Name{Francesca L. Crowe}
      \Email{f.crowe@bham.ac.uk}\\ 
       \addr Institute of Applied Health Research, University of Birmingham
       \AND
       \Name{Tom Marshall}
       \Email{t.p.marshall@bham.ac.uk}\\ 
       \addr Institute of Applied Health Research, University of Birmingham
      \AND
  \Name{Sylvia Richardson}
       \Email{sylvia.richardson@mrc-bsu.cam.ac.uk}\\ 
       \addr MRC Biostatistics Unit, University of Cambridge
         \AND
  \Name{Paul D. W. Kirk}
  \Email{paul.kirk@mrc-bsu.cam.ac.uk}\\ 
       \addr MRC Biostatistics Unit, University of Cambridge}

\begin{document}
\maketitle
\begin{abstract}

    We present a one-shot, unsupervised federated learning approach for Bayesian model-based clustering of large-scale binary and categorical datasets, motivated by the need to identify patient clusters in privacy-sensitive electronic health record (EHR) data. We introduce a principled `divide-and-conquer’ inference procedure using variational inference with local merge and delete moves within batches of the data in parallel, followed by `global' merge moves across batches to find global clustering structures. We show that these merge moves require only summaries of the data in each batch, enabling federated learning across local nodes without requiring the full dataset to be shared.  Empirical results on simulated and benchmark datasets demonstrate that our method performs well relative to comparator clustering algorithms. We validate the practical utility of the method by applying it to a large-scale British primary care EHR dataset to identify clusters of individuals with common patterns of co-occurring conditions (multimorbidity). 
\end{abstract}

\paragraph*{Data and Code Availability}
Code is available \href{https://github.com/j-ackierao/FedMerDel-code}{here}, which includes code for generation of all simulated datasets and random seeds used in this paper.

The MNIST dataset used is publicly available (\url{http://yann.lecun.com/exdb/mnist/}), and details and code for preprocessing are available in the above repository. The THIN EHR dataset used is not publicly available, but we provide details regarding preprocessing in the Appendix.

\paragraph*{Institutional Review Board (IRB)}
Collection of data in THIN was approved by the NHS South East Multi-Centre Research Ethics Committee (MREC). Approval to conduct this analysis was obtained from the Scientific Review Committee (reference number: 21SRC055).

\vspace{-\baselineskip}

\section{Introduction}
\label{sec:intro}
Federated learning (FL) refers to the setting in which a network of clients collaboratively train a model while keeping training data local \citep{Kairouz2021}, introduced for privacy-sensitive datasets \citep{McMahan2017} common in healthcare. While supervised FL has seen many developments, unsupervised FL methods remain limited \citep{Tian2024}. With the primary motivation of clustering electronic health record (EHR) data across multiple hospitals, we address: (i) preserving data privacy while learning global clustering structure, (ii) enabling scalability to large patient record datasets, and (iii) maintaining clustering accuracy in distributed environments. While various federated clustering algorithms and Bayesian FL methods exist \citep{IJCAI2023}, there is limited work on FL for unsupervised Bayesian model-based clustering.
Our method implements a one-shot FL scheme where clients perform local computation and share summary statistics with a central server in a single round, enabling asynchronous participation ideal for healthcare institutions. By combining clustering information across sites, we identify population-level disease clusters transcending institutional boundaries, enabling characterization of multimorbidity patterns across diverse populations impossible with site-specific clustering alone.

Mixture models provide a probabilistic approach for clustering by assuming data are generated from a mixture of underlying probability distributions. Bayesian mixture models enable the uncertainty in cluster allocations and, potentially, the number of clusters to be modelled. 
Markov Chain Monte Carlo (MCMC), particularly Gibbs sampling \citep{Neal2000}, is often used to draw samples from the intractable posterior \citep[e.g.][]{Robert2004, Walker2007}. Despite efforts to improve scalability, MCMC algorithms are often too computationally intensive for even moderately large datasets, and/or may experience mixing issues, preventing the sampler from reaching convergence \citep{Celeux2000}.  

To address the growth of `massive' datasets, a range of methods have been proposed to scale algorithms to large sample sizes \citep[e.g][]{Huang2005}. These include distributed computing approaches in which the computational and memory requirements of an algorithm are split, with the algorithm running in parallel on multiple nodes \citep{Zhu2017}. However, models with unidentifiable latent variables present additional complexity for such approaches. Furthermore, few distributed computing approaches are in the spirit of FL, where data heterogeneity, data privacy, asynchronous client participation and node dropout are challenges to address. 

While \textcolor{black}{variational inference (VI)} methods tend to have better run-time performance than MCMC algorithms \citep[e.g.][]{Blei2017}, 
they can also be sensitive to initialisation due to local optima in the objective function. To address these issues when fitting mixture models, `merge' and `delete' moves may be incorporated into the variational algorithm to dynamically combine or remove unnecessary clusters. Such moves, in addition to `split' moves to create new clusters, have improved inference in expectation-maximisation (EM) algorithms \citep{ueda_smem_2000} and MCMC samplers \citep{jain_split-merge_2004, Dahl2021} for mixture models, and have also been employed less commonly in VI algorithms for Dirichlet process (DP) mixtures, including DP mixtures of Gaussians \citep{pmlr-v38-hughes15, NIPS2013_8c19f571}.

Here we propose a `one-shot' distributed algorithm that performs inference for Bayesian mixture models using VI with local merge and delete moves within batches of the data across a network of nodes in parallel. This is followed by `global' merge moves to aggregate clusters across batches, utilising the variational objective function to combine results from data partitions in a statistically principled manner. 




We focus on applications to categorical, particularly binary (2-category), datasets, as is common in EHR datasets such as the one studied later in the manuscript. \textcolor{black}{Binary data analysis has a long history in the context of characterising multimorbidity across populations \citep[see e.g.][]{Cornell2009-wb, Ng2018-sl, Nichols2022-hq} but there has been limited research into clustering methodology for binary and categorical data.} By applying our method to large-scale EHR data, we demonstrate how global clustering structures (in this case, \textcolor{black}{population-level} multimorbidity patterns) can be identified without compromising data privacy. 
Our approach can be adapted to other members of the exponential family, and we consider in the Appendix how our model can be extended to tackle variable selection. 

\section{Related Work}

A central challenge in scaling and distributing Bayesian inference is maintaining statistical accuracy while distributing computation. One could consider parallelising steps within VI to enhance scalability, and we look at this in Section \ref{paralellisedmerdel}. In the MCMC setting, an important example of distributed inference is \emph{consensus Monte Carlo} \citep{Scott2016}, where Monte Carlo sampling is performed independently on partitions of the data, and the results are then aggregated to approximate a global posterior. While effective for certain models, these approaches typically struggle with models involving unidentifiable latent variables. Informally, the challenge for clustering models is how to match clusters/cluster labels across data partitions. \cite{pmlr-v37-gea15} considers a distributed inference algorithm for the DP and HDP mixture models with a variant of the DP slice sampler, but requires frequent communication between servers. The SNOB \citep{Zuanetti2018} and SIGN \citep{Ni2019} algorithms also perform inference for distributed mixture models, where MCMC samplers are run locally on partitions of the data, before local clusters are combined into global clusters with another sampler, allowing for asynchronous operation. These models are conceptually closest to our proposed model, although they do not consider the federated learning (FL) setting. 

Distributed VI algorithms have also been proposed to obtain an approximation of the posterior distribution for mixture models, including variational consensus Monte Carlo (VCMC) \citep{NIPS2015_e94550c9}, where the authors suggested a heuristic approach using the Hungarian algorithm \citep{Kuhn1955} to align cluster centres across data partitions. Our proposed method instead utilises the variational objective function to combine results from data partitions in a statistically principled manner. \cite{NIPS2015_38af8613} proposes a streaming, distributed VI method for the DP mixture model, but does not address data privacy. \cite{pmlr-v89-zhang19c} extends stochastic VI \citep{JMLR:v14:hoffman13a} to a distributed setting for the inference of mixture models, but relies on frequent parameter exchange between processors, despite efforts to make the algorithm asynchronous. 

The majority of work on Bayesian inference in FL targets supervised tasks \citep[see][for a survey]{IJCAI2023}. These approaches have been extended to tackle e.g. Bayesian neural networks for local classification or regression tasks \citep{pmlr-v222-bhatt24a}, and Bayesian nonparametric methods for dynamic learning of local models \citep{NEURIPS2021_46d0671d}. Some works explore clustering clients in FL to tackle data heterogeneity in supervised settings \citep{Sattler2021, Briggs2020}. This is popular in personalised FL, concerning the learning of individual local distributions for clients in a federated network. For example, in \cite{Zhang2023}, clients in the same cluster share a prior, and \cite{pmlr-v202-wu23z} represents clients with mixtures of shared-parameter distributions where clients have different mixture weights. These methods aim to improve performance on downstream prediction tasks, but do not cluster data itself, differing fundamentally from our objective. 

Far fewer works have explored \textit{unsupervised} FL. Most adopt heuristic approaches like federated $k$-means \citep{Kumar2020,Liu2020,Garst2025, pmlr-v139-dennis21a} or fuzzy clustering \citep{Stallman2022, Pedrycz2022} lacking probabilistic foundations, while FedEM \citep{NEURIPS2021_f740c8d9} provides probabilistic modeling but requires frequent communication rounds.

Overall, our work fills a gap at the intersection of unsupervised FL, Bayesian clustering, and scalable distributed inference. To the best of our knowledge, we are the first to propose a communication-efficient, one-shot federated algorithm for Bayesian model-based clustering of categorical and binary data.


\section{Model Overview}

\subsection{Bayesian Finite Mixture Models} \label{bayesianmixmodels}

Let ${X} = \{{x}_1, ..., {x}_N\}$ denote the observed data, where ${x}_n$ is one observation of $P$ categorical variables. We model the generating distribution as a finite mixture of $K$ components. Each observation is generated by one component. The general equation for a $K$ component finite mixture model is:

{\footnotesize
\begin{align}
&p(X|\boldsymbol{\pi}, \boldsymbol{\phi})= \prod_{n=1}^N p(x_n | \boldsymbol{\pi}, \boldsymbol{\phi}),
\end{align}
\begin{align}
\qquad &\mbox{where} \qquad p(x_n | \boldsymbol{\pi}, \boldsymbol{\phi}) = \sum_{k=1}^K \pi_k f(x_n | \phi_k),\label{mixture}
\end{align}}
where component densities $f(x_n | \phi_k)$ are usually from the same parametric family but with different parameters associated with each component. Under certain assumptions, the Dirichlet process (DP) mixture model may be derived by considering the limit as $K$ goes to infinity \citep{Maceachern1994, Escobar1995, Neal2000}. We consider the so-called `overfitted mixtures' setting in which $K$ is set to be larger than the number of clusters we expect, which enables the number of clusters to be inferred \citep{Rousseau2011}. Each component is modelled as a categorical distribution across $P$ covariates with parameters $\phi_{kj} = [\phi_{kj1}, ..., \phi_{kjL_j}]$, where $\phi_{kjl}$ represents the probability that variable $j$ takes value $l$ in component $k$. $L_j$ is the number of categories for variable $j$. The mixture weight for the $k$-th component is denoted by $\pi_k$, satisfying $\sum_{k=1}^K \pi_k = 1$ and $\pi_k \geq 0$.  

We introduce latent variables $z_n$ associated with each $x_n$, such that each $z_n$ is a one-hot-encoded binary vector with $z_{nk} = 1$ if and only if $x_n$ is in the $k$-th cluster. We then rewrite Equation \eqref{mixture} as:

\vspace{-0.6em}
{\footnotesize \begin{equation}
p(X|Z, \boldsymbol{\pi}, \boldsymbol{\phi}) = \prod_{n=1}^N\prod_{k=1}^K f({x}_n | \phi_{k})^{z_{nk}}.
\end{equation}}

To complete the model specification for the Bayesian model, we place priors on the parameters:

\vspace{-0.6em}
{\footnotesize \begin{align}
    \boldsymbol{\pi} = (\pi_1, ..., \pi_K) &\sim \mbox{Dirichlet}(\alpha_0), \label{piprior} \\
    \phi_{kj} = (\phi_{kj1}, ..., \phi_{kjL_j}) &\sim \mbox{Dirichlet}(\epsilon_{j}), \label{phiprior} 
\end{align}}
where all Dirichlet priors are symmetric. It has been shown theoretically that if $K$ exceeds the true number of clusters, then, under certain assumptions, setting $\alpha_0 < 1$ in Equation \eqref{piprior} allows the posterior to asymptotically converge to the correct number of clusters as the number of observations goes to infinity \citep{Rousseau2011, vanHavre2015, MalsinerWalli2014}. Further details of hyperparameter settings are provided in the Appendix. The model can be extended to simultaneously tackle variable selection, as in \cite{Law2004} and \cite{Tadesse2005}, which is explored in the Appendix. \textcolor{black}{This accounts for noisy features not contributing to the clustering structure, which can help address data corruption in real-world data.}

\subsection{Mean-Field Variational Inference (VI)}

We employ a variational inference (VI) approach to seek \textcolor{black}{an approximate distribution $q$ close to the true posterior}. We optimise the Evidence Lower Bound (ELBO) below, $\mathcal{L}(q)$, with respect to $q(\theta)$, where $\theta$ is a collection of all parameters in the model. 

{\footnotesize
\begin{equation}
    \mathcal{L}(q) = \int q(\theta) \ln \left(\frac{p(X, \theta)}{q(\theta)} \right)d\theta
\end{equation}}

This is equivalent to minimising the Kullback-Leibler (KL) divergence between $q(\theta)$ and the true posterior, $p(\theta | X)$. We constrain $q$ to be a mean-field approximation which can be fully factorised as: $q(\theta) = q_Z(Z)q_\pi(\pi)q_\Phi(\Phi)$. We optimise $\mathcal{L}(q)$ using a standard variational EM algorithm, \textcolor{black}{also known as Coordinate Ascent Variational Inference} \citep[see, for example,][]{Bishop2006}, as described below. 

\paragraph{`E' Step}

We update the latent variables $Z$ with the current variational distributions of all other parameters.

{\footnotesize
\begin{align}
q^\ast(Z) = \prod_{n=1}^N\prod_{k=1}^K r_{nk}^{z_{nk}}, \qquad r_{nk} = \frac{\rho_{nk}}{\sum_{j = 1}^K \rho_{nj}} \label{respexplanation} \\
    \ln \rho_{nk} = \mathbb{E}_{ \pi}[\ln {\pi}_k] + \sum_{i=1}^P \mathbb{E}_{\Phi}[\ln \phi_{kix_{ni}}] \label{rhodef}
\end{align}}

$r_{nk}$ is the responsibility of the $k$-th component for the $n$-th observation, and $\mathbb{E} [z_{nk}] = r_{nk}$. 

\paragraph{`M' Step}

Given updated responsibilities $r_{nk}$ in the `E' step, we then update cluster-specific parameters:

{\footnotesize
\begin{align}
    q^\ast(\pi) &= \mbox{Dirichlet}(\alpha^\ast_1, ..., \alpha^\ast_K) \label{pidef1}, \\ \qquad q^\ast(\phi) &= \prod_{k=1}^{K} \prod_{j=1}^{P} \mbox{Dirichlet}(\epsilon^\ast_{kj1}, ..., \epsilon^\ast_{kjL_j})
\end{align}}
where for $k = 1, ..., K, j = 1, ..., P, l = 1, ..., L_j$:
{\footnotesize 
\begin{align}
    \alpha^\ast_k &= \alpha_0 + \sum_{n=1}^N r_{nk}, \qquad \epsilon^\ast_{kjl} = \epsilon_{jl} + \sum_{n=1}^N \mathbb{I}(x_{nj} = l)r_{nk} \label{epsdef}
\end{align}}

Expectations throughout are taken with respect to variational distributions, with ELBO monitored for convergence. All values are initialised using k-modes \citep{Chaturvedi2001}. \textcolor{black}{The algorithm terminates when the ELBO is within a certain tolerance for 3 iterations in a row.} 

\section{Variational Merge/Delete Moves} \label{compcons}

As in e.g. \citet{pmlr-v38-hughes15}, we introduce merge and delete moves which can be integrated into the VI algorithm. Merge moves combine observations from two clusters into one unified cluster; delete moves remove unnecessary extra clusters. Both moves effectively eliminate redundant clusters to help the model to escape local optima, and are designed in a principled manner to align with the ELBO variational objective function. When moves are proposed, parameters in the VI framework are updated through initially updating responsibilities, $r_{nk}$ which are directly associated with cluster assignment. \textcolor{black}{For example, to merge clusters $k_1$ and $k_2$, we set $r_{nk^\ast} = r_{nk_1} + r_{nk_2}$ for a new unified cluster $k^\ast$, and perform a variational M step to update $\pi, \phi$. We perform additional E and M steps to allow observation reassignment, and accept moves only if they improve the ELBO, otherwise we maintain the original configuration.} 

Variational implementations of these moves in Dirichlet process mixture models, in addition to split moves, have been shown to outperform standard variational Bayes and related algorithms \citep{pmlr-v38-hughes15, Yang2019}. However, little work has been published regarding variational categorical mixture models. 
Clear, practical strategies for the implementation of these moves in the categorical setting are provided in Appendix \ref{sec:mergedeleteappend}. \textcolor{black}{One of our key contributions lies in developing efficient strategies specifically for categorical mixture models to select candidate clusters to merge/delete.} Comparing ELBOs between all possible merge/delete proposals would result in a high computational cost. We demonstrate in Appendix \ref{sec:corrdivrand} that using fast-to-calculate heuristics such as correlation between cluster parameters \textcolor{black}{or divergence between variational distributions when proposing to merge} allows for enhanced efficiency. This aspect has rarely been explored.

We name the variational algorithm including merge/delete moves `MerDel' and introduce a parameter, `laps', representing how many variational EM cycles we go through before performing a merge and a delete move. Since merge and delete moves include additional E and M steps (see Appendix \ref{sec:mergedeleteappend}), they are costly compared to standard variational steps which guarantee ELBO improvement. 
Subsequently, there is a trade-off between proposing more merge/delete moves to allow for bigger, more effective moves earlier on in MerDel versus reducing wasted computation as the model approaches convergence and moves are rejected, which we examine in simulations. One method to mitigate this in future could be to dynamically change the `laps' parameter. 

\subsection{Parallelising MerDel} \label{paralellisedmerdel}

VI requires fewer sweeps over the dataset to perform inference compared to MCMC samplers, and is feasible for datasets with thousands of observations. This allows variational mixture models to be more computationally efficient, especially with merge/delete moves. However, VI still reaches computational bottlenecks when $N$ is very large; we see in simulations later that MerDel is infeasible as we reach datasets with $N$ of order $10^5$. Repeatedly updating $N \times K$ responsibilities every E step, and subsequently summing these over $N$ every M step is costly. 

Individual components of the original algorithm can be parallelised using standard methods e.g., in the E step, calculating $r_{nk}$ can be parallelised over $n$. 
We provide an implementation of such a parallelised VI as part of this manuscript. However, the overall algorithm is not embarrassingly parallel; for example, $r_{nk}$ values for all $N$ observations are required to update cluster-specific parameters for a given cluster $k$. Prior literature has consistently highlighted the limited gains achieved by such partial parallelisation for iterative methods such as VI. This is due to the significant computational overhead incurred when feeding intermediate results to and from one `master' core - a globally shared computation unit - at frequent intervals \citep{Nallapati2007, neiswanger2015embarrassinglyparallelvariationalinference}. 

\section{Federated Variational Inference}


In this section, we propose a scalable clustering algorithm for large categorical datasets, FedMerDel (Federated MerDel), which  addresses the challenge of large $N$ and enables federated learning (FL). 

In FedMerDel, we use a `divide and conquer' inference procedure. We assume our dataset has been split into $B$ `batches' - these batches will already be pre-determined if subsets of the data are held by separate parties, or may be artificially created by random partitioning of the data. The first `local' step of the algorithm involves carrying out clustering via MerDel in each data batch separately. The size and number of these batches should be chosen so inference is computationally feasible; pre-existing subsets of the data can be divided further if necessary. Each batch is processed in parallel, and the clusters produced are referred to as `local clusters'. 

The second `global' step then combines the local clustering structures. Local clusters are frozen, and we merge similar local clusters into `global' clusters. This is conceptually similar to approaches used in SNOB \citep{Zuanetti2018} and SIGN \citep{Ni2019}, but both use MCMC methods which are computationally costly; local clustering in SNOB takes over 4 hours for a batch of size $N=1000$ in the authors' simulations. We utilise a novel version of the merge move in MerDel to implement a `global merge'. By freezing local clusters in the global merge, the proposed clustering structure and `global ELBO' across all observations can be efficiently calculated by utilising stored parameter values. The variational framework allows us to accept or reject global merges in a principled manner. Importantly, the full dataset is not required in the global merge.

In contrast to the approach considered in Section \ref{paralellisedmerdel}, instead of optimising individual components of the existing classical variational algorithm, our approach is designed to present a scalable clustering model which avoids these bottlenecks entirely by adopting a distributed computing approach in which each data batch is analysed independently and in parallel before `global merge' steps are used to obtain a (global) mixture model for the full dataset. This FL approach eliminates the need for frequent communication and synchronization during the main variational updates, while data remains decentralised. 

\subsection{Estimating Global Clusters}

Assume our data has been split into $B$ batches and MerDel has been run on each batch, with $K_b$ local clusters in Batch $b$. Let $K = \sum_{b=1}^B K_b$ be the total number of local clusters. \textcolor{black}{We begin with $\alpha^\ast$ and $\epsilon^\ast$ parameters fixed from the completed local clustering steps.} In the global step, we first `combine' all our existing variational parameters from each batch. For $q(\pi)$, we concatenate all our $K_b$-length $\alpha^\ast$ vectors to one $K$-length $\alpha^\ast$ vector. Each $q(\phi_{kj})$ is already independent for each $(k, j)$ pair. We create a new matrix of responsibilities $r_{nk}$ as an $N \times K$ block diagonal matrix with blocks of $N_b \times K_b$ responsibility matrices from each batch. $\sum_{k=1}^K r_{nk} = 1$ still holds for all $n$.

Let clusters $k_1$ and $k_2$ be considered for a merge. As in the local merge, we reassign responsibilities to cluster $k_1$ via $r^{new}_{nk_1} = r_{nk_1} + r_{nk_2}$ for all $n = 1, ..., N$. This can be viewed as summing columns $k_1$ and $k_2$ in our responsibility matrix. We then use the definitions of $\alpha^\ast$, $\epsilon^\ast$ (Equation \eqref{epsdef}) as follows to update cluster-specific parameters:

{\footnotesize
\begin{align}
    \alpha^{\ast new}_{k_1} &= \alpha_0 + \sum_{n=1}^N ( r_{nk_1} + r_{nk_2} ) = \alpha^\ast_{k_1} + \alpha^\ast_{k_2} - \alpha_0 \\
    \epsilon^{\ast new}_{k_1jl} &= \epsilon_{jl} + \sum_{n=1}^N \mathbb{I}(x_{nj} = l) (r_{nk_1} + r_{nk_2}) = \epsilon^\ast_{k_1jl} + \epsilon^\ast_{k_2jl} - \epsilon_{j}
\end{align}
}
This avoids calculating sums over $N$. To complete the merge, we remove all parameters associated with cluster $k_2$ - $\alpha^\ast_{k_2}$, $\epsilon^\ast_{k_2j}$ and $r_{nk_2}$ for all $n = 1, ..., N$, $j = 1, ..., P$. Unlike local merge moves, we do not perform additional E/M steps to allow observations to move in and out of the new cluster. This would require a whole-dataset update of responsibilities $r_{nk}$. The efficiency of the global merge hinges on local clusters being frozen. We propose global merge candidates using two approaches: greedy search and random search (see Appendix \ref{description:randomgreedy} for details). 

\subsection{Efficient Global ELBO Calculation} \label{sec:efficientELBO}

The ELBO variational objective function allows us to accept or reject the global merge under a rigorous probabilistic framework. For the global merge to be accepted, the ELBO must be higher. As we now consider the full dataset, many terms in the overall ELBO function involve all observations and clusters. `Freezing' local clusters allows us to efficiently calculate the majority of the ELBO terms and enable the model to handle large-scale data. Two sums involving $N$ in the full ELBO function can be written as a sum over smaller $K$, $P$, $L_j$ of sufficient statistics which can in turn be calculated as a linear function of previously calculated $\alpha^\ast$ and $\epsilon^\ast$: $T_k = \sum_{n=1}^N r_{nk} = \alpha^\ast_k - \alpha_0$ and $S_{kjl} = \sum_{n=1}^N r_{nk} \mathbb{I}(x_{nj} = l) = \epsilon^\ast_{kjl} - \epsilon_{jl}$. This vastly reduces the number of operations required.

The final term involving $N$ is an assignment entropy term, $\sum_{n=1}^N \sum_{k=1}^K r_{nk} \ln r_{nk}$, which is not linear in any previously calculated parameters. This is additive over $N$ and $K$, so we can calculate this value separately for each batch (in parallel if required) before summing results. However, in our greedy search, we never combine two clusters from the same batch, so $r_{nk_1}$ and/or $r_{nk_2}$ will \textit{always} be zero for any merge, for all $n = 1, ..., N$. Observation $n$ is only explained by clusters originating from some batches of data and a merge is only proposed from a new batch where $n$ is not explained by any clusters associated with that batch. This allows greedy search to be faster, although merges could be missed. See Appendix \ref{sec:globalelbo} for full details of the global ELBO calculation.

Crucially, the global merge move allows us to seek a clustering structure for the full dataset without any scan over the full data matrix $X$. Data from separate entities remains decentralised, where only summary statistics from each batch are required to be shared, allowing for increased data security and privacy. \textcolor{black}{These summary statistics represent cluster-specific weighted averages over batch-level populations, and do not directly expose individual-level data. We note that privacy risks could occur with clusters where these weights are almost all estimated to be close to 0 with a small number close to 1, but such cases can be easily detected and handled (eg. by withholding the cluster).} Implementation details are in Appendix \ref{appendix:implementation}. Parallelisation in local batches (Section \ref{paralellisedmerdel}) could be applied to FedMerDel for additional speed gains.


\section{Experiments and Results} 

An additional study looking at unsupervised clustering of the MNIST dataset \citep{MNIST} can be found in Appendix \ref{appendix:mnist}.

\subsection{Simulation Study}

In this section, simulated data are binary, with the data generating mechanism detailed in the Appendix. Additional simulation studies comparing merge criteria, looking at categorical data with 3+ categories, and examining variable selection are in the Appendix. 

\subsubsection{Frequency of Merge/Delete Moves}

In Section \ref{compcons}, we described the trade-off between more or fewer merge/delete moves in MerDel. We ran simulation studies comparing different values of the parameter `laps' representing the frequency of merge/delete moves in a run of MerDel, with laps $\in \{1, 2, 5, 10\}$. We also compared to a model with just merge/delete moves after performing one `E' and `M' step after initialisation (`laps = 0'), and the variational model with no merge/delete moves (`laps = 10000').  We compared the wall-clock time taken for convergence, number of non-empty clusters found, and the Adjusted Rand Index (ARI) with the simulated `true' labels. Appendix \ref{setup:freqmerdel} gives further details.

We saw that models incorporating merge/delete moves (laps~$\in\{2,5,10\}$) consistently outperformed standard variational inference across multiple simulated datasets, achieving higher ARIs, more accurate cluster number estimation, and faster convergence (full results in Appendix \ref{suppfreqresults}). Based on these results, we used laps~$=5$ for subsequent experiments.

\subsubsection{Global Merge Simulations} \label{mainglobalresults}

We evaluated the clustering performance when splitting data into batches and performing a global merge as detailed in Section 5. In our simulations, we compared to the ``gold standard" setting in which the full dataset may be analysed without \textcolor{black}{splitting of the data} using MerDel with or without parallelisation (denoted by `par' and `full' in our results tables). We looked at binary data with sizes varying between $N =$ between 10,000 to 1,000,000 with $P$ = 100 covariates. 

For our simulated datasets, we generated data with known ground truth cluster structure, then initially randomly partitioned this into batches. We also looked at scenarios with heterogeneous data across partitions, \textcolor{black}{where clusters are deliberately generated such that they may only appear in certain batches and not in others.} We compared this to parallelised VI on the full data and compared to existing unsupervised federated and distributed algorithms such as k-FED \citep{pmlr-v139-dennis21a}, FedEM \citep{NEURIPS2021_f740c8d9} and SNOB \citep{Zuanetti2018}. These represent the current state-of-the-art for unsupervised federated clustering, particularly for categorical data, in this nascent field. We also compared to some centralised learning methods as well as simulations varying $P$ and looking at categorical data in the Appendix. Full details are in Appendix \ref{setup:globalmerge}.

A full results table can be found in Appendix \ref{globalmergeappendix}, with results for heterogeneously distributed data shown in Table~\ref{globalmergeresults}. 

\begin{table*}[!hbtp]
\floatconts
 {globalmergeresults}
 {\caption{A subset of results for `Global Merge Simulations', where data is distributed heterogeneously. For (a), (b), (c), the heterogeneous data scenarios are described in Appendix \ref{setup:globalmerge}. We report the median and lower/upper quantiles across all 10 `shuffles'/initialisations and all 10 synthetic datasets. FedMerDel results use random search.}}
 {\footnotesize
 \renewcommand{\arraystretch}{0.9}
 \begin{tabular}{lccccr}
\toprule
\bfseries Scenario & \bfseries Model & \bfseries Batches/Cores & \bfseries ARI & \bfseries Clusters & \bfseries Time (s) \\
\midrule
(a) & par & 10 & 0.945 [0.944, 0.951] & 12 [12, 12] & 119 [110, 131]\\  
 & FedMerDel & 10 & 0.942 [0.936, 0.946] & 13 [12, 13] & 64.5 [61.5, 70.1] \\
\midrule
(b) & par & 5 & 0.955 [0.949, 0.956] & 10 [10, 10] & 169 [161, 192] \\
& FedMerDel & 5 & 0.993 [0.992, 0.994] & 10 [10, 10] & 131 [127, 165] \\
\midrule
(c) & par & 5 & 0.950 [0.944, 0.950] & 12 [12, 12.8] & 69.5 [65.8, 77.3] \\
& FedMerDel & 5 & 0.988 [0.985, 0.990] & 13 [12, 13] & 53.2 [48.3, 58.3] \\
\bottomrule
 \end{tabular}}
\end{table*}

When the data were divided into random batches, FedMerDel achieved clustering performance nearly equivalent to the ``best possible'' centralised baselines, with median ARIs only marginally below those of the full and parallel models (all above 0.9). While local batch-level ARIs were also high, the global merge step provides additional value by producing a unified model for the entire population. In heterogeneous scenarios, FedMerDel occasionally surpassed the full-data baseline (ARI~$\approx$~0.99). This occurs when batches contain fewer, well-separated clusters, making local clustering more accurate and improving the global merge outcome. Thus, FedMerDel accommodates statistical heterogeneity without forcing clusters to merge across batches, and can offer clustering-quality advantages beyond computational efficiency.

The time taken was substantially faster for both parallelised MerDel and FedMerDel as expected via distributed computing, where MerDel on the full dataset became infeasible as our data sizes became larger due to the computational requirements of manipulating large matrices. However, as hypothesised in Section \ref{paralellisedmerdel}, we saw that FedMerDel was faster than parallelised MerDel, and became substantially more so as $N$ increased (Figure \ref{Nlines}). 

\begin{figure}[htbp]
\floatconts
  {Nlines}
  {\caption{Plot comparing the mean time taken by MerDel with and without parallelisation (`par' and `full'), and FedMerDel as we vary $N$ with 5 or 10 batches/cores in `Global Merge Simulations', with an approximate 95\% confidence interval (mean $\pm$ 1.96 $\times \frac{\text{s.d}}{\sqrt{n}}$). FedMerDel results use random search.}}
  {%
    \subfigure[]{\label{glob5}%
      \includegraphics[width=0.49\linewidth]{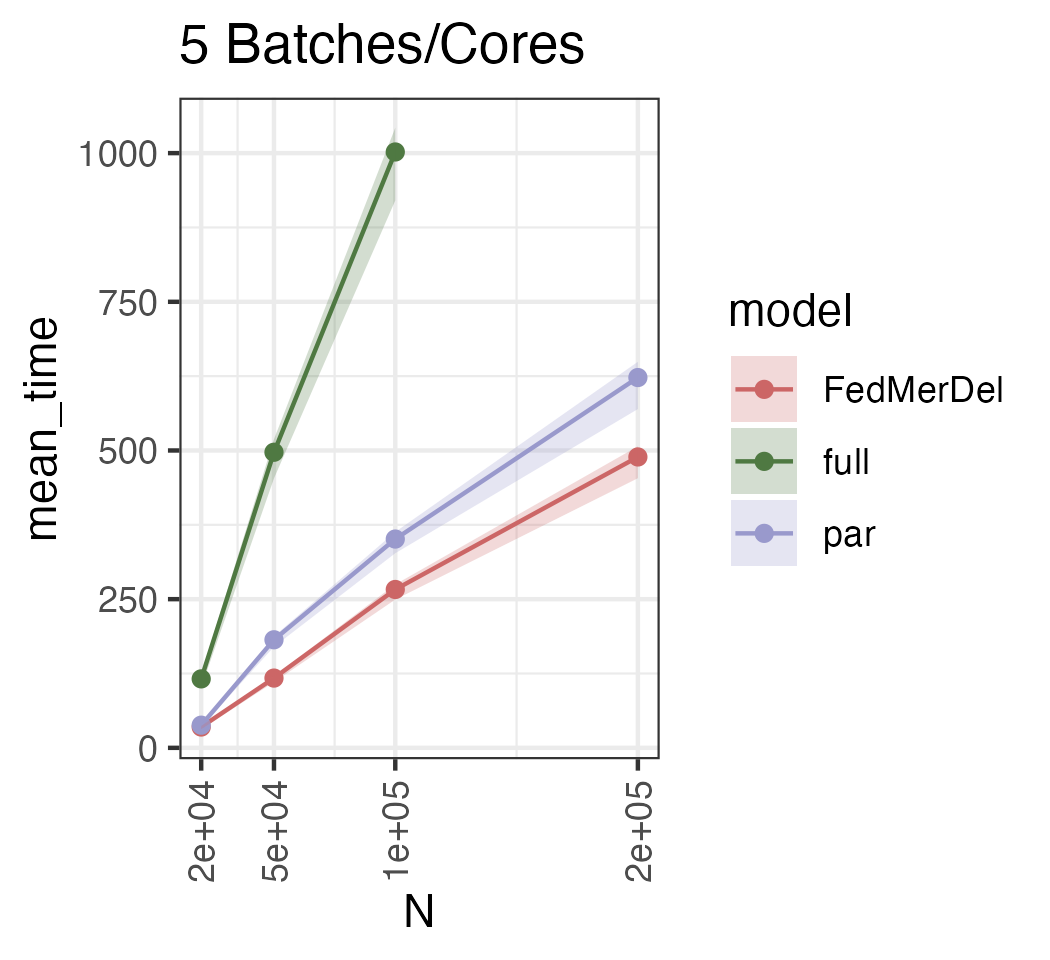}}%
    \subfigure[]{\label{glob10}%
      \includegraphics[width=0.49\linewidth]{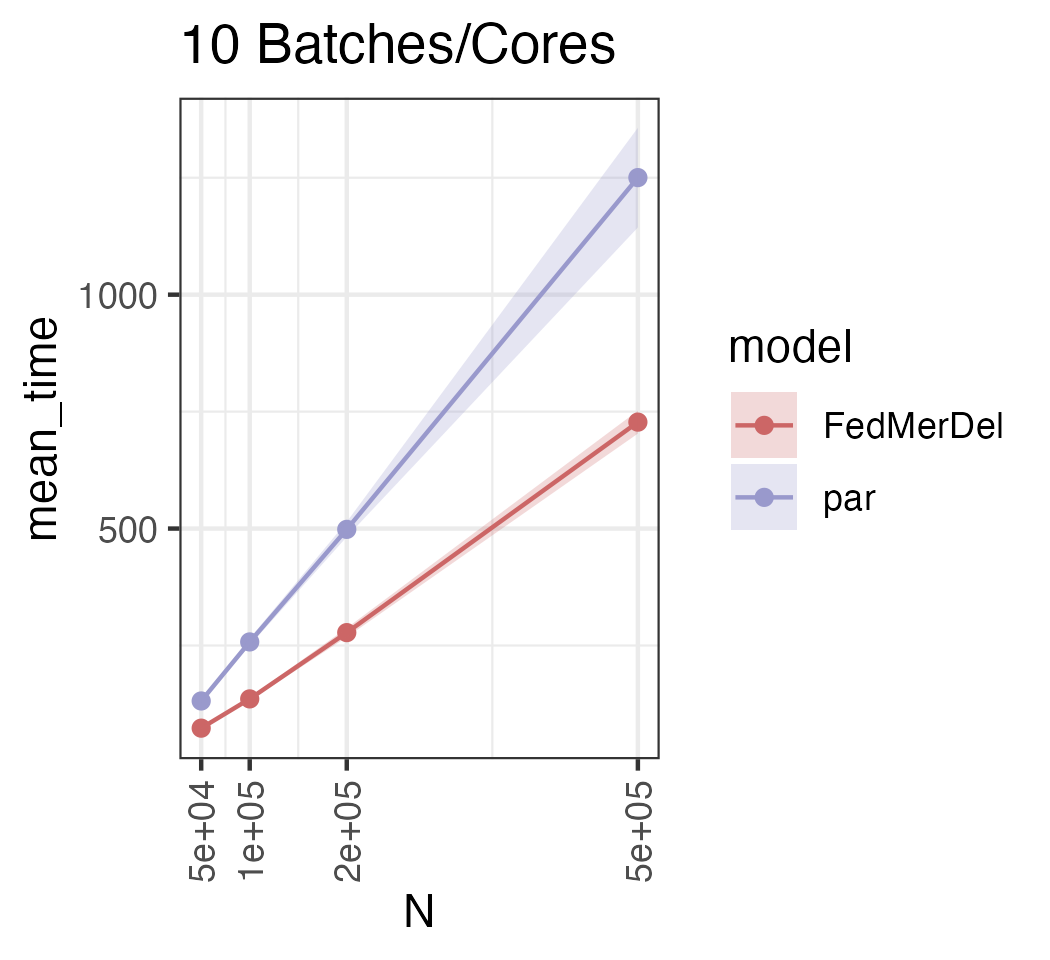}}
  }
\end{figure}

Global merge times ranged from a mean of 0.29s(greedy)/0.39s(random) for $N=20,000$ to a mean of 14.8s(greedy)/18.2s(random) for $N=500,000$, showing that the time taken for the global merge was not computationally prohibitive. \textcolor{black}{Different batches of data for a dataset also provided consistent results, e.g. the standard deviation of ARIs across 10 different shuffles of data in an $N = 50,000$ simulation ranged from 0.0003 to 0.005 for 20 datasets.}

FedMerDel outperformed all comparison methods in terms of both accuracy and time efficiency. k-FED performed poorly (ARI between 0-0.2). FedEM involved frequent communication between clients and a central server, and took 1.5-4 hours per model. Local Gibbs sampling in parallel took 20-40 mins; FedMerDel took $<$3 mins (Table~\ref{comparisontable}). We also compared FedMerDel to non-federated clustering approaches including k-modes, hierarchical clustering, DBSCAN, and latent class analysis (LCA; Appendix~\ref{appendix:unsupervisedcomp}). While the heuristic algorithms were computationally efficient, they produced near-zero ARIs on our simulated datasets, indicating poor recovery of true clustering structures.  LCA performed better (mean ARI $\approx$  0.87 at $N = 50,000$) but required multiple model fits and longer runtimes ($\sim$10 min per model).

\begin{table*}[htbp]
\floatconts
  {comparisontable}
  {\caption{ARI results comparing unsupervised FL methods. k-FED is omitted due to poor performance. Where $N$ is stated, observations were assigned to partitions randomly; `heterogeneous (c)' is described in Appendix \ref{setup:globalmerge}. `Cl'=clients, `ba'=batches.}}
  {\footnotesize
  \renewcommand{\arraystretch}{0.9}
  \begin{tabular}{lccccr}
\toprule
\bfseries Sim & \bfseries FedEM & \bfseries Approx SNOB & \bfseries FedMerDel \\
\midrule
N=20000 & 10 cl: 0.847 [0.798, 0.865]& 5 ba: 0.699 [0.695, 0.700] & 5 ba: 0.920 [0.912, 0.933] \\
& 20 cl: 0.906 [0.849, 0.913] & & \\
\midrule
N=50000 & 25 cl: 0.839 [0.793, 0.879] & 10 ba: 0.836 [0.814, 0.861] & 10 ba: 0.951 [0.946, 0.954] \\
\midrule
Heterogeneous (c) & 10 cl: 0.907 [0.902, 0.913] & 5 ba: 0.502 [0.439, 0.632] & 5 ba: 0.988 [0.985, 0.990] \\
\bottomrule
  \end{tabular}}
\end{table*}

\subsubsection{Number of Batches} 
With datasets of size $N=100,000$ and $N=200,000$, we compared the clustering performance when considering different numbers of batches (further details in Appendix~\ref{setup:noofshards}). As we increased the number of batches used, although there was no statistically significant difference at the 5\% level in the clustering accuracy as measured by ARI, we did see more clusters (Appendix~\ref{result:numberofbatches}). ARI values were all above 0.93.

\subsection{Electronic Health Record (EHR) data}
We analysed an anonymised EHR dataset derived from The Health Improvement Network (THIN) database that comprised \textcolor{black}{primary-care} records from 289,821 individuals in the United Kingdom over the age of 80. For 94 long-term health conditions, we recorded which conditions were present in each individual to create a $289,821 \times 93$ binary matrix whose $(i,j)$-entry indicates whether individual $i$ has condition $j$ (encoded as a 1) or not (encoded as 0). Further details on data preprocessing are provided in Appendix \ref{EHRappendix}. The aim of the analysis is to identify clusters of individuals with common patterns of co-occurring conditions (multimorbidity) across a population, allowing us to discover clinically meaningful subgroups that support population health research.

We divided the data into 14 batches of size 20,000 and 1 batch of size 9,821 using random partitioning, and ran FedMerDel with $K = 20$ maximum clusters within each batch.  After global merging, we obtained 12 clusters, which are summarised in Figure \ref{THINsummary}. Each cluster can be characterised by particular conditions.  From top to bottom in Figure \ref{THINsummary}, we have clusters of individuals with: (i) cancer; (ii) Atrial fibrilation (AF) \& arrhythmia; (iii) peripheral vascular disease \& aortic aneurysm; (iv) stroke; (v) AF \& arrhythmia and stroke; (vi) stroke and blindness; (vii) blindness; (viii) dementia; (ix) asthma and chronic obstructive pulmonary disease; (x)-(xii) ``generally multimorbid" individuals.

To characterise these clusters and quantitatively validate their clinical relevance, we examined the mean number of conditions per person in clusters and cluster-level mortality statistics, \textcolor{black}{which serve as an objective clinical outcome measure (see Appendix~\ref{sec:randomEHRchar}). We found that our multimorbidity clusters were associated with different mortality levels, providing quantitative evidence that FedMerDel identified clinically relevant patient subgroups.} 
For example, the stroke + AF/arrhythmia cluster has higher mortality than the stroke-only cluster, highlighting potential differences in clinical risk profiles. Of the ``generally multimorbid" clusters, (x) sees a lower mean number of conditions per person and a lower mortality rate compared to the population average, (xi) is associated with psychiatric disorders and lower mortality, while cluster (xii) represents a high-mortality multimorbid group. While there may be further subclusters in cluster (x), we believe these do not emerge due to weak signals compared to core clusters. Future work using recursive clustering approaches could enable the discovery of such finer-grained structure.

\begin{figure}[htbp]
\floatconts
  {THINsummary}
  {\caption{EHR clustering results with random search. Columns are health conditions, and rows correspond to clusters. Shading indicates proportion of individuals in each cluster with each condition; row labels show the number of individuals per cluster. To improve visualisation, only the most prevalent health conditions are shown. See Appendix~\ref{EHRappendix} for further details, coding of health conditions and greedy search results.}}
  {\includegraphics[width=\linewidth]{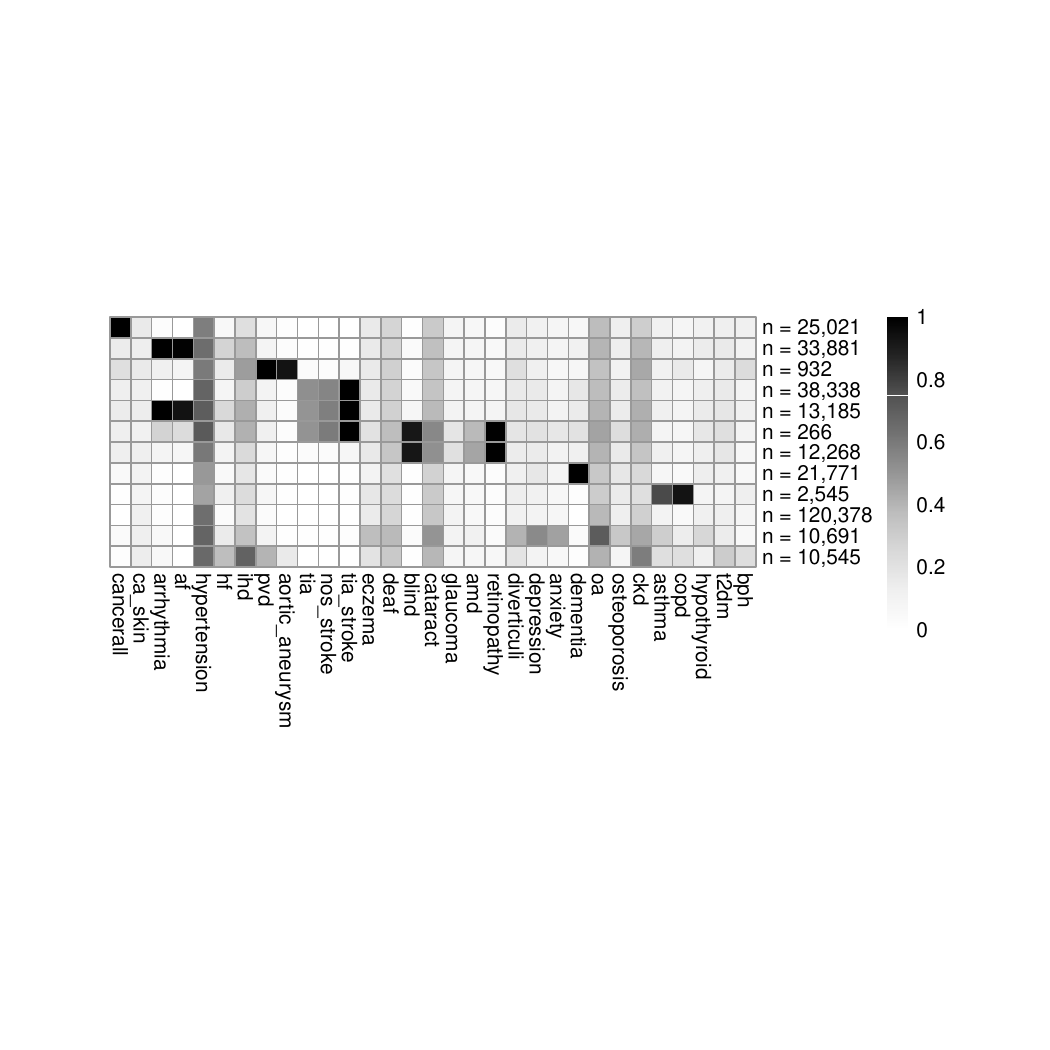}}
\end{figure}

To evaluate performance under realistic heterogeneity \textcolor{black}{expected when combining data from different healthcare sites}, we simulated heterogeneity in the THIN dataset (details and full results in Appendix \ref{sec:hetehr}). When dementia patients were separated into distinct batches, the method revealed subclusters within the dementia group, supporting future work into recursive clustering. Using a Dirichlet distribution to simulate condition-based skews across batches, core clusters were retained, while smaller subclusters emerged, likely driven by batch-specific over-representation. \textcolor{black}{Beyond computational efficiency, this federated approach offers valuable insights into cluster prevalence in batches - in real EHR settings, such patterns could identify region-specific multimorbidity clusters based on different sociodemographic characteristics of local populations. FedMerDel recovers meaningful population-level patterns even in the presence of inter-node variability.}

\section{Discussion} 

In this manuscript, we introduced FedMerDel, a Bayesian mixture modelling algorithm using variational inference and `global' merge moves to enable federated learning (FL) without sharing the full dataset across nodes. This model showed good performance on both simulated and real-world data, with only slight reductions in performance compared to centralised analysis, and excelled when data was not identically distributed across batches, a key challenge in FL. FedMerDel outperformed other existing unsupervised FL methods, where it could directly address both the challenge of scaling model-based clustering, and the challenge of clustering decentralised data. This is valuable for scenarios requiring both data privacy and scalability e.g. patient health records. 

There is clearly scope for improvement in some aspects of the model. \textcolor{black}{While we push scalability beyond other probabilistic approaches (eg. MCMC, EM), physical hardware limits remain, with memory use from large parameter sets creating computational bottlenecks. To scale further,} we could add extra steps in the global merge as in the SIGN model \citep{Ni2019}, where local clusters are combined over multiple steps over multiple computing nodes or cores. \textcolor{black}{Efficiency could be improved by adapting the `laps' parameter controlling the frequency of merge/delete moves as MerDel approaches convergence, similar to learning rate scheduling.} Future work could explore allowing local cluster splits during global merging (\cite{Song2020-qb} represents local clusters as a mixture of distributions) and theoretical convergence guarantees for global merge moves. \textcolor{black}{Further VI techniques including Stochastic Variational Inference (SVI) \citep{JMLR:v14:hoffman13a} and variational tempering \citep{pmlr-v51-mandt16} could also be implemented within each data batch to further improve scalability and accuracy of the local clustering.} 

While we presented our approach in the context of modelling large categorical datasets, we note that it can be straightforwardly adapted to mixtures of Gaussians and other members of the exponential family for other applications. The use of sufficient statistics to allow for efficient ELBO calculation could extend to any exponential family distribution; \cite{hughes_memoized_2013} provides further details. 

\acks{}

The project was part of the Bringing Innovative Research Methods to Clustering Analysis of Multimorbidity (BIRM-CAM) project that was funded by the MRC (MR/S027602/1). We also acknowledge MC\_UU\_00040/05.

\bibliography{mergedelete}

\newpage
\appendix

\onecolumn

\section{Variable Selection}

As explored in \citet{Law2004} and \citet{Tadesse2005}, we can introduce binary feature selection indicators $\gamma_j$ - feature saliencies - where $\gamma_j = 1$ if and only if the $j$-th covariate is relevant to the clustering structure, and irrelevant variables have their feature saliencies reduced to zero. The probability density for a data point $x_n$ in cluster $k$ is given by:
\begin{align}
f({\mathbf{x}}_n|\Phi_k) &= \prod_{j = 1}^P f_j(x_{nj} | \Phi_{kj})^{\gamma_j} f_j(x_{nj} | \Phi_{0j})^{1 - \gamma_j}
\end{align}

$\Phi_{0j} = [\phi_{0j1}, ..., \phi_{0jL_j}]$ are parameter estimates obtained for covariate $j$ under the assumption that there exists no clustering structure in the $j$-th covariate. These are precomputed using maximum likelihood estimates as seen in \citet{Savage2013}. The priors associated with $\gamma_j$ for $j=1, ..., P$ are as follows:

\begin{equation}
\gamma_j | \delta_j \sim \mbox{Bernoulli}(\delta_j)
\end{equation}
\begin{equation}
    \delta_j \sim \mbox{Beta}(a, a)
\end{equation}

\textcolor{black}{These irrelevant variables can correspond to noisy/corrupted features, which is particularly useful in real-world cases where we believe there may be data corruption.}



\section{Variational Updates with Variable Selection} \label{variableselectionupdates}

\paragraph{`E' Step}
This follows a similar form to the model with no variable selection, where $Z$ takes a multinomial distribution.
\begin{equation}
q^\ast(Z) = \prod_{n=1}^N\prod_{k=1}^K r_{nk}^{z_{nk}}, \qquad r_{nk} = \frac{\rho_{nk}}{\sum_{j = 1}^K \rho_{nj}} \label{respexplanation2}
\end{equation}
\begin{align}
    \ln \rho_{nk} &= \mathbb{E}_{ \pi}[\ln {\pi}_k] + \sum_{j=1}^P c_j \mathbb{E}_{\Phi}[\ln \phi_{kjx_{nj}}] + (1 - c_j)(\ln \phi_{0jx_{nj}}) \label{rhodef2}
\end{align}
$c_j = \mathbb{E}_\gamma(\gamma_j)$, where the expectation is taken over the variational distribution for $\gamma$.

\paragraph{`M' Step}

Similarly to the model with no variable selection, the updates for the cluster-specific parameters are given by:
\begin{align}
    q^\ast(\pi) &= \mbox{Dirichlet}(\alpha^\ast_1, ..., \alpha^\ast_K) \\
    q^\ast(\phi) &= \prod_{k=1}^{K} \prod_{j=1}^{P} \mbox{Dirichlet}(\epsilon^\ast_{kj1}, ..., \epsilon^\ast_{kjL_j})
\end{align} 
where for $k = 1, ..., K, j = 1, ..., P, l = 1, ..., L_j$:
\begin{align}
    \alpha^\ast_k &= \alpha_k + \sum_{n=1}^N r_{nk} , \qquad \epsilon^\ast_{kjl} = \epsilon_{j} + \sum_{n=1}^N \mathbb{I}(x_{nj} = l)r_{nk}c_j
\end{align}

The updates for $\gamma$ and $\delta$ are given as:
\begin{equation}
    q^\ast(\gamma_j) = \mbox{Bernoulli}(c_j), \qquad c_j = \frac{\eta_{1j}}{\eta_{1j} + \eta_{2j}} = \mathbb{E}_\gamma(\gamma_j)
\end{equation}
\begin{align}
    \ln \eta_{1i} &= \sum_{n=1}^N \sum_{k=1}^K (r_{nk} \mathbb{E}_\Phi[\ln \phi_{kjx_{nj}}]) + \mathbb{E}_\delta [\ln \delta_j] \\
    \ln \eta_{2i} &= \sum_{n=1}^N \sum_{k=1}^K (r_{nk} \ln \phi_{0jx_{nj}}) + \mathbb{E}_\delta [\ln (1 - \delta_j)]
\end{align}
\begin{align}
    q^\ast(\delta_j) = \mbox{Beta}(c_j + a, 1 - c_j + a)
\end{align}

As before, all expectations throughout are taken over the variational distributions for each parameter and we alternate between `E' and `M' steps.

\section{Merge and Delete Moves: Further Detail} \label{sec:mergedeleteappend}

\subsection{Merge Moves} 
Merge moves refer to moves that combine observations from two clusters into one unified cluster. Suppose clusters $k_1$ and $k_2$ are considered for a merge. We propose a move to reassign all observations from clusters $k_1$ and $k_2$ into a single new cluster, $k^\ast$. The responsibilities for this new cluster are given by $r_{nk^\ast} = r_{nk_1} + r_{nk_2}$. All other responsibilities for other clusters remain the same. 

We then perform a `dummy' variational M step, updating the $\pi, \phi$ parameters. We run an extra E and an extra M step to allow observations to move in or out of the new cluster. We calculate the ELBO and compare this to the ELBO before the merge move; we accept this move if the ELBO has improved, otherwise we return to the model prior to the merge. If accepted, we reassign cluster $k^\ast$ to the $k_1$ position and delete cluster $k_2$ from the model. 

\subsubsection{Selecting Candidate Clusters to Merge} \label{howtomerge}

We could choose a pair of clusters for a merge at random, but a merge is more likely to be accepted if the two clusters are `similar'. Making informed decisions on which pair of clusters to propose for a merge can improve efficiency by reducing the number of rejected merge moves. Two methods of assessing similarity between clusters using correlation and divergences between variational distributions are detailed below. Other ways include comparing marginal likelihoods or posterior probabilities of candidate models as in \citet{pmlr-v38-hughes15} and \citet{ueda_smem_2000}. 

Due to the stochastic nature of the selection from the `candidate clusters', similar results between different merge candidates are to be expected and the choice of measure is insignificant. Results from a simulation study comparing selection criteria for clusters to be merged are seen in Appendix \ref{sec:corrdivrand}. Correlation and divergence-based methods performed similarly, but did provide a reduction in overall computation time and an improvement in rates of accepted merges compared to random selection. 

\paragraph{Correlation}

We can use an assessment of correlation between the variational parameters associated with clusters; a notion of correlation is seen also in \citet{hughes_memoized_2013}.

We construct a correlation score by looking at the correlation between the parameter values $\epsilon^\ast$ in the distributions $q(\phi_{kj})$. In the binary case, each $\epsilon^\ast_{kj}$ is given by a two dimensional vector. For each k, we can take the first value of each of these vectors for $j = 1, ..., P$, concatenate these into one $j$ length vector, $\epsilon^{corr}_{k}$ and then compare the correlation of these concatenated vectors across clusters. The $\epsilon^{corr}_k$ values represent the (expected) frequency of 0's in each cluster across all variables in a binary dataset of 0's and 1's. 

We select randomly between the 3 cluster pairs with the highest correlation, provided $\mbox{Corr} (\epsilon^{corr}_{k_1}, \epsilon^{corr}_{k_2})$ is above 0.05. With the categorical case with $L$ categories per variable, we could consider finding the correlation scores for $L-1$ concatenated vectors across all variables for $L-1$ of the categories and taking the mean correlation score. However, this is complicated when variables have different numbers of categories.

\paragraph{Divergence-based Measures}

We also look at using measures of distance between probability distributions to quantify cluster similarity. For a pair of clusters $k_1$ and $k_2$, we look at the divergence between $q(\phi_{k_1j})$ and $q(\phi_{k_2j})$ and sum over all variables $j$ for a pair of clusters $k_1$ and $k_2$. We then eg. choose randomly between the 3 cluster pairs with the lowest divergences. Many divergences could be considered; in this paper we considered the Kullback-Leibler (KL) divergence and the Bhattacharyya distance. We emphasise that we are not seeking a precise characterisation of the `distance' but are only making approximate comparisons.

\subsection{Delete Moves}

Delete moves refer to moves to delete unnecessary extra clusters in the model. Redundant small clusters often remain at local optima in variational algorithms for mixture models. 

Let cluster $k$ be a candidate cluster for deletion. For the set of observations in cluster $k$, ie. $S_k := \{n = 1 \ldots N: z_{nk} = 1\}$, we create a new data-frame $X_{-k}$ with the rows from observations in $S_k$ removed, and the column for cluster $k$ removed. We first perform the variational E step after removing all variational parameters relevant to cluster $k$ in the model. This recalculates all responsibilities for the remaining observations. We then perform the variational M step, which updates cluster-specific parameters.

We then return to the original dataframe $X$ and then run an E step using $X$ with the current model parameters, which reassigns all observations in $X$ to new clusters. After another M step, we then calculate the value of the ELBO function. We accept the delete move if the new ELBO is higher.

Note that in both the merge and delete moves described above, there is a small approximation involved in setting $r_{nk} = 0$ exactly for responsibilities associated with clusters `removed'. Appendix \ref{sec:priors} gives further detail; this approximation simply replaces infinitesimally small numbers (smaller than $10^{-80}$) with 0.

\subsubsection{Selecting Candidate Clusters to Delete}

Smaller clusters are likely to be better candidates for deletion as it is likely these observations have been `left out'. Extremely small clusters are also less informative in practical applications and may be less reproducible in other similar independent datasets. In this paper, we propose a cluster chosen randomly between all clusters which are smaller than 5\% of the dataset; if there are none, we choose randomly between the three smallest clusters. 

\section{Global Merge - Further Details} 

\subsection{Random and Greedy Search} \label{description:randomgreedy}

\paragraph{Greedy Search}

We prevent merges between two clusters from the same batch with a greedy search, on the hypothesis that such merges would already have taken place at the local stage. This can be carried out in a sequential manner by examining the correlation/divergences between Cluster 1 of Batch 1 and all clusters in Batch 2. Once Cluster~1 has merged with any cluster in Batch~2, we move to Batch 3 and consider merging with the clusters in this batch. We repeat for all batches, before moving to Cluster 2 of Batch 1 and so on until we have gone through all clusters in Batch 1. We then repeat the process with Batches 2, 3, ..., B, looking only at Batches $b + 1, ..., B$ for Batch $b$.

\paragraph{Random Search}

Similarly to local merge moves, we calculate correlations/divergences between all $K$ clusters, and pick the merge candidate pair randomly between the 3 pairs with the highest correlation (provided correlation $>$ 0.05) or lowest divergence. There is no restriction on proposing merges between two clusters of the same batch. This stops after a fixed number of global merge moves (e.g. 10) have been rejected consecutively. 

\subsection{Global ELBO Calculation}\label{sec:globalelbo}

For the model with no variable selection, the ELBO for the full dataset is given by:
\begin{align*}
    \mathcal{L}(q) &= \mathbb{E}_{Z, \pi, \Phi}[\ln p(X, Z, \pi, \Phi)] - \mathbb{E}_{Z, \pi, \Phi}[\ln q(Z, \pi, \Phi)] \\ 
    &= \mathbb{E}_{Z, \Phi}[\ln p(X | Z, \Phi)] + \mathbb{E}_{Z, \pi}[\ln p(Z|\pi)] + \mathbb{E}_{\pi}[\ln p(\pi)] + \mathbb{E}_{\Phi}[\ln p(\Phi)] \numberthis \\
    & \qquad - \mathbb{E}_{Z}[\ln q(Z)] - \mathbb{E}_{\pi}[\ln q(\pi)] - \mathbb{E}_{\Phi}[\ln q(\Phi)] \label{ELBObreakdown}
\end{align*}
where all expectations are taken with respect to the variational distributions for $Z, \pi, \Phi$. 
We use the current $\alpha^\ast$ and $\epsilon^\ast$ values to calculate the values of sufficient statistics:
\begin{align}
T_k &= \sum_{n=1}^N r_{nk} = \alpha^\ast_k - \alpha_0 \\
S_{kjl} &= \sum_{n=1}^N r_{nk} \mathbb{I}(x_{nj} = l) = \epsilon^\ast_{kjl} - \epsilon_{jl}
\end{align}

Each term in Equation \eqref{ELBObreakdown} is as follows:
\begin{align}
    \mathbb{E}_{Z, \Phi}[\ln p(X | Z, \Phi)] &= \mathbb{E}_{Z, \Phi}[\sum_{n=1}^N \sum_{k=1}^K {z_{nk}} (\sum_{j=1}^P \ln \phi_{kjx_{nj}})] = \sum_{k=1}^K \sum_{j=1}^P \sum_{l = 1}^{L_j} S_{kjl} \mathbb{E}_{\Phi}[\ln \phi_{kjl}] \\
    \mathbb{E}_{Z, \pi}[\ln p(Z|\pi)] &= \mathbb{E}_{Z, \pi} [\sum_{n=1}^N \sum_{k=1}^K {z_{nk}} \ln \pi_k] = \sum_{k=1}^K T_k \mathbb{E}_{\pi} [\ln \pi_k] \\
    \mathbb{E}_{\pi}[\ln p(\pi)] &= -\ln B(\alpha) + \sum_{k=1}^K (\alpha_k - 1) \mathbb{E}_{\pi}[\ln \pi_k] \label{ELBOppi} \\
    \mathbb{E}_{\Phi}[\ln p(\Phi)] &= \sum_{k=1}^K \sum_{j=1}^P (- \ln B(\epsilon_{kj}) + \sum_{l=1}^{L_j} (\epsilon_{kjl} - 1) \mathbb{E}_{\Phi}[\ln \phi_{kjl}]) \label{ELBOpphi} \\
    \mathbb{E}_{Z}[\ln q(Z)] &= \sum_{n=1}^N\sum_{k=1}^K r_{nk} \ln r_{nk} \qquad \mbox{(assignment entropy)} \label{ELBOqz} \\
    \mathbb{E}_{\pi}[\ln q(\pi)] &= -\ln B(\alpha^\ast) + \sum_{k=1}^K (\alpha^\ast_k - 1) \mathbb{E}_{\pi}[\ln \pi_k]  \\
    \mathbb{E}_{\Phi}[\ln q(\Phi)] &= \sum_{k=1}^K \sum_{j=1}^P (- \ln B(\epsilon^\ast_{kj}) + \sum_{l=1}^{L_j} (\epsilon^\ast_{kjl} - 1) \mathbb{E}_{\Phi}[\ln \phi_{kjl}]) \label{ELBOqphi}
\end{align}

Apart from the assignment entropy term in Equation \eqref{ELBOqz}, as we have calculated $T_k$ and $S_{kjl}$ via simple linear operations on our model parameters, there are no sums over $N$ and all terms are mainly sums over much smaller $K$, $P$, $L_j$. Furthermore, some terms in $B(\alpha)$, $B(\alpha^\ast)$, $B(\epsilon_{kj})$, $B(\epsilon^\ast_{kj})$ which cancel out. Importantly, $X$ appears nowhere in the calculations; our global merge move allows us to seek a global clustering structure without any full scan over the $N \times P$ matrix $X$. 


\subsection{Global Merge - Variable Selection} \label{description:varsel}

When finding a global clustering structure in the model with variable selection, our focus is on updating cluster allocation via merging local clusters, so we use the same global ELBO function as in the model without variable selection and only update $Z, \pi, \phi$. 
Implementing this amounts to a small approximation in the true ELBO for the true Bayesian model. We consider a variable to be `selected' based on the percentage of times a variable was selected across the batches. In later simulations, we consider a variable selected if it was selected in either all batches, or all but one batch. This borrows ideas of `thresholds' for variable selection across multiple VI runs, seen in \citet{Rao2024}.

\section{Approximations and Priors in MerDel} \label{sec:priors}

In Appendix \ref{sec:mergedeleteappend}, we detailed that in the process of performing our merge/delete moves, we `remove' the clusters by eliminating their parameters and responsibilities from the model code. This could raise concerns pertaining to transdimensional inference. However, we implicitly account for the removed clusters in subsequent calculations. We call these removed clusters `zombie clusters'.

This is because the cluster-specific parameters $\alpha, \epsilon$ are determined by the prior only (as there is no data assigned to the cluster to update cluster-specific parameters). We can accurately calculate ELBO values and track these zombie clusters throughout the algorithm. The key approximation is setting responsibilities for zombie clusters to exactly 0. These values are not usually precisely 0, but are infinitesimally small.

\subsection{Emptying a Cluster + M Update} \label{sec:M}

When a cluster $k^\ast$ is removed (through merge or delete), this is equivalent to assigning $r_{nk^\ast} = 0$ for all observations $n$, effectively zeroing column $k^\ast$ in the $N \times K_{\mbox{init}}$ responsibility matrix. The subsequent variational M step updates $\alpha^\ast, \epsilon^\ast$:
\begin{equation}
    \alpha^\ast_k = \alpha_0 + \sum_{n=1}^N r_{nk} , \qquad \epsilon^\ast_{kjl} = \epsilon_{jl} + \sum_{n=1}^N \mathbb{I}(x_{nj} = l)r_{nk}  \label{alphaandeps}
\end{equation}

As we have that $r_{nk} = 0$ for all $n = 1, ..., N$ for a removed cluster $k$, $\alpha^\ast_k = \alpha_0$ and $\epsilon^\ast_{kjl} = \epsilon_{jl}$. These variational posteriors for cluster $k$ are solely ruled by the priors for $\pi$ and $\Phi$.

\subsection{The Approximation: E Update} \label{sec:Eupdate}

In MerDel, we then perform a variational E update with the full dataset to allow observations to move between clusters after cluster-specific parameter updates. Recall that responsibilities $r_{nk}$ are defined by:
\begin{equation}
r_{nk} = \frac{\rho_{nk}}{\sum_{j = 1}^K \rho_{nj}}, \qquad \ln \rho_{nk} = \mathbb{E}_{ \pi}[\ln {\pi}_k] + \sum_{j=1}^P \mathbb{E}_{\Phi}[\ln \phi_{kjx_{nj}}] \label{eqn:respandrho}
\end{equation}

The expectations in $\ln \rho_{nk}$ are:
\begin{align}
    \mathbb{E}_{ \pi}[\ln {\pi}_k] &= \psi(\alpha^*_k) - \psi(\sum_{k = 1}^K \alpha^*_k) \label{eqn:Epi} \\
    \mathbb{E}_{\Phi}[\ln \phi_{kjx_{nj}}] &= \psi(\epsilon^*_{kjx_{nj}}) - \psi(\sum_{l = 1}^{L_j} \epsilon^*_{kjl}) \label{eqn:Ephi}
\end{align}

Since, for $x > 0$, the digamma function $\psi(x)$ increases approximately exponentially, observations are assigned to clusters with responsibilities close to 1. These clusters therefore have $\alpha^\ast_k$ values on the scale of the number of observations assigned - sometimes in the thousands - which are much larger than $\alpha_0 < 1$. Therefore $\ln \rho_{nk}$ is much smaller for `zombie clusters', resulting in $r_{nk}$ values \textit{extremely} close to 0.

These near-zero $r_{nk}$ values persist as we continue through the EM steps; $\alpha^\ast$ and $\epsilon^\ast$ still remain extremely small for removed clusters (virtually equal to prior values) in comparison to clusters with observations, and $r_{nk}$ values still remain virtually 0. Our approximation sets $r_{nk} = 0$ permanently by removing these columns, improving computational efficiency and memory usage. Since cluster-specific parameters for removed clusters equal their priors when $r_{nk}=0$, we also remove these.

This approximation has virtually no effect on our model; we run two small simulations comparing the effects of implementing this approximation, and saw that true $r_{nk}$ values were often on the scale of $10^{-70}$ to $10^{-100}$ on typical MerDel datasets. After an accepted MerDel proposal, we continue with standard variational EM steps where removed clusters remain permanently eliminated.

This behaviour mirrors the overfitted variational model without merge/delete, where empty clusters naturally maintain near-zero responsibilities and virtually never regain observations. ELBO calculation remains straightforward; terms from zombie clusters either cancel each other out or are trivially calculated.

\subsection{Approximations and Priors in FedMerDel}

There is no approximation when considering a `global merge' in FedMerDel, as we never perform any variational E or M steps after performing merges. Responsibilities $r_{nk}$ remain exactly 0 for `zombie clusters'. 

In the global setting, we define a new Dirichlet prior for $\pi$ over our full dataset with the total number of local clusters initialised: $B \times K_b$, where $K_b$ is the initialised number of clusters in Batch $b$. This is justified as each cluster has a prior where:
\begin{equation}
    p(\pi) \propto \prod_{k=1}^K {\pi_k}^{\alpha_0 - 1}
\end{equation} 
with K representing the maximum desired clusters per batch. The maximum clusters globally should be $B \times K_b$, the maximum total clusters when we consider all batches. A fractional prior for each batch would inappropriately constrain the model to $K_b$ clusters across the entire dataset, despite searching $K_b$ clusters per partition. We do not want to constrain the model to force all clusters to merge across batches; in FL, there are often clusters only appearing in certain batches. This is consistent with literature; the same Dirichlet priors for mixing probabilities are used in `local clusters' and `global clusters' in \citep{Zuanetti2018}, a similar MCMC model.


\section{Implementation Details} \label{appendix:implementation}
The model builds upon an existing R package, VICatMix \citep{Rao2024}, which applies VI for binary and categorical Bayesian finite mixture models (with no merge/delete moves). Rcpp and RcppArmadillo are used to accelerate computation with C++. MerDel runs on a single-core processor, where, for example, inference for a mixture model fitted to a binary dataset of size $N=2000$, $P=100$ generally converges in less than a minute.

FedMerDel allows MerDel to be run independently on a different core per batch of data; reported `wall-clock times' take this parallelisation into account where clearly stated. 

\section{Simulation Set-Up - Additional Information}

\subsection{Data Generating Mechanism}
We simulated synthetic binary data with $N$ observations and $P$ covariates by sampling the probability $p$ of a `1' in each cluster for each variable via a $\mbox{Beta} (1, 5)$ distribution, encouraging sparse probabilities. For noisy variables, the probability of a `1' was also generated by a $\mbox{Beta} (1,5)$ distribution but this probability was the same for every observation regardless of cluster membership. When simulating categorical data with $L$ categories, we instead used a $\mbox{Dirichlet} (1, ..., L)$ distribution.

\subsection{Hyperparameter Choice}
In all simulations (real-world and simulated data), we set $\alpha_0 = 0.01$, $\epsilon_j = 1/L_j$ and $a = 2$ (as in \citet{Rao2024}) in the priors in Section 2 of the main paper. We did not tune these in this manuscript as all examples are purely illustrative. All algorithms were run with a maximum of 1000 iterations (an iteration being one variational EM step) - although this limit was not reached for any simulation. 

Convergence tolerance was set between 0.000005 and 0.00000005 in all simulations. When comparing wall-clock time for convergence, all simulations in a given study had the same convergence tolerance. In all simulations involving FedMerDel, we used the parameters `laps = 5' for MerDel runs, except when specified. 

\subsection{Frequency of Merge/Delete Moves} \label{setup:freqmerdel}

Details of the scenarios for this simulation study are given in Table \ref{simFedMerDeltable}. Simulations 1.1, 1.2, 1.3 were run without variable selection; Simulations 1.4, 1.5 were run with variable selection. $K_{\mbox{init}}$ refers to the initialised value of $K$ in the model. In each scenario, we generated 20 independent datasets and compared 10 different initialisations for each dataset. We used correlation for merge criteria.

\begin{table*}[h!]
\caption{Table giving parameters for data generation for the first simulation study, `Frequency of Merge/Delete Moves'.}
\label{simFedMerDeltable}
\begin{center}
\begin{footnotesize}
\begin{sc}
\begin{tabular}{lcccccr}
\toprule
ID & Relevant & $N$ & $P$ & $K_{init}$ & $K_{true}$ & $N$ per \\
 & Variables &  &  &  &  & Cluster \\
\midrule
    1.1 &  100 & 1000 & 60 & 20 & 5 & 100-300\\
    1.2 &  100 & 2000 & 100 & 20 & 8 & 50-800 \\
    1.3 &  100 & 4000 & 100 & 25 & 10 & 200-800 \\
    1.4 &  75 & 1000 & 100 & 20 & 10 & 50-200 \\
    1.5 & 60 & 1500 & 100 & 20 & 10 & 50-200 \\
\bottomrule
\end{tabular}
\end{sc}
\end{footnotesize}
\end{center}
\end{table*}

\subsection{Global Merge Simulations} \label{setup:globalmerge}
We aimed to evaluate the clustering performance of FedMerDel when splitting a simulated dataset into equally sized batches, and compared this to a parallelised version of MerDel (on the full dataset) as well as the usual MerDel algorithm on the full dataset (variational EM with merge and delete moves). We set `laps = 5', $P=100$ covariates, 12 true clusters and 20 initialised clusters in all cases. Correlation was used for all merge criteria. Parameters were kept the same to enable comparison of run-times between different values of $N$. Details of the values of $N$ tested and number of batches/cores used are given in Table \ref{globalmergesetuptable}. 

The full, unparallelised algorithm was only used for the first 4 simulations listed, as it became computationally infeasible for larger values of $N$ with our hardware. 

\begin{table*}[h!]
\caption{Table giving parameters for data generation for the second simulation study, `Global Merge Simulations', where other parameters are set as in Section \ref{setup:globalmerge}.}
\label{globalmergesetuptable}
\begin{center}
\begin{footnotesize}
\begin{sc}
\begin{tabular}{lr}
\toprule
N & Batches/Cores\\
\midrule
20000 & 5\\
50000 & 5\\
50000 & 10 \\
100000 & 5\\
100000 & 10 \\
200000 & 5\\
200000 & 10 \\
500000 & 10 \\
\bottomrule
\end{tabular}
\end{sc}
\end{footnotesize}
\end{center}
\end{table*}

\paragraph{Heterogeneous Data Simulations}  As the above simulations have observations assigned purely randomly to batches, we also look at situations with heterogeneous (non i.i.d) data in batches, as is common in FL. We consider scenarios where a) 1 of 12 clusters is only seen in one batch and no others with $N=50,000$, b) 10 clusters are equally split between 5 batches with $N=50,000$ and c) 10 clusters are equally split between 5 batches, with 2 more clusters which are evenly split across the 5 batches with $N=20,000$. In all cases, we set $P = 100$. We compared this to parallelised VI on the full data. 

\paragraph{Comparisons to Other Unsupervised FL Methods} We compared to other unsupervised FL methods for simulated data, including FedEM for a categorical mixture model \citep{NEURIPS2021_f740c8d9}, and an adaptation of k-FED \citep{pmlr-v139-dennis21a} that uses k-modes instead of k-means as appropriate for categorical data. We also compared an approximate version of SNOB \citep{Zuanetti2018}, with local clusters found via Gibbs sampling (with R package PReMiuM \citep{Liverani2015}), then merged using hierarchical clustering. We looked at randomly distributed simulated data with $N=20,000$ and $N=50,0000$, and also looked at heterogeneous data scenario c) in the above paragraph.

\paragraph{Comparisons to Other Unsupervised Centralised Learning Methods} \textcolor{black}{To provide additional context for MerDel and FedMerDel's clustering performance, we compared MerDel and FedMerDel to centralized clustering methods suitable for large binary datasets on simulated datasets of size $N=20,000$ and $N=50,000$. Methods we looked at were k-modes \citep{Chaturvedi2001} (recall this is also used for initialisation of MerDel), hierarchical clustering with `hclust' in R \citep{Murtagh2011}, and DBSCAN \citep{dbscan} - in all cases, we fed the algorithm the true number of clusters, 12. We also evaluated latent class analysis (LCA), which previously performed well on similar binary EHR data \citep{Nichols2022-hq} using the poLCA R package \citep{Linzer2011}.}


\paragraph{Varying $\mathbf{P}$}
We additionally compared the ARI, number of clusters and the time taken for clustering models using parallelised MerDel and FedMerDel when we varied $P$. In this simulation, we simulated 5 synthetic datasets and ran MerDel and FedMerDel with 5 `shuffles'/initialisations each time. Datasets were all of size $N = 100,000$ with 10 equally sized clusters, and we initialised with 20 clusters in all cases. We split into 5 equally sized batches for FedMerDel, and parallelised over 5 cores for parallelised MerDel. 

\paragraph{Less Separable Data}
\textcolor{black}{To address the challenge where clusters are less easily separated, we also ran simulations where the probability of a ‘1’ was generated by a Beta(2, 2) distribution. Further, we simulated data with this generating distribution and where two clusters - clusters 11 and 12 - had 75\% of their covariates with their cluster-specific probabilities being $\pm$ 0.1 of each other. In both of these simulations, we set $N=50,000$, $P=100$ and 12 true clusters, initialised with 15 clusters. 5 independent datasets were generated, and FedMerDel was run 3 times on each.}

\paragraph{Global Merge with Variable Selection}

For this simulation with variable selection (on binary data), we employed the `greedy search' for the global merge, and used correlation to assess similarity between clusters. The two scenarios had simulated data split into 5 equally sized batches with $N =$ 20,000, 50,000, with $K = $ 15 and 10 equally sized true clusters respectively. We set $P = 100$ covariates in both cases, where 80\% were relevant to the clustering structure in the first simulation and 75\% were relevant in the second simulation. We generated 5 simulated datasets and had 5 different `shuffles' of the data, as well as running MerDel on the full data 5 times (unparallelised and parallelised), as with other `global merge' simulations. 

As well as looking at ARI, number of resulting clusters and wall-clock time as with other simulations, we looked at the number of selected variables. As described in Section \ref{description:varsel}, for FedMerDel, we consider a variable selected if it was selected in either all batches, or all but one batch (where a variable is selected if $c_j > 0.5$, the expectation of the variable selection latent variable (Appendix \ref{variableselectionupdates})).

\subsection{Number of Batches} \label{setup:noofshards}

With datasets of size $N=100,000$ and $N=200,000$, we compared the clustering performance when considering $B=2, 4, 10$ and $B=4, 8, 20$ batches respectively. Batches were sized equally. We simulated 20 different synthetic datasets with $P=100$ for each scenario and took 5 different `shuffles' of the data for every global merge model. Models had 10 unevenly sized `true' clusters (cluster sizes ranging from 5\% to 20\% of the dataset) and were initialised with 20 clusters.

\section{Additional Simulation Results}

\subsection{Frequency of Merge/Delete Moves} \label{suppfreqresults}



\textbf{Incorporating merge/delete enables faster convergence in terms of wall-clock time: } Generally, MerDel models are faster to converge. Figure \ref{Sim1linegraphs} in Appendix \ref{suppfreqresults} shows that early accepted merge/delete moves allow for substantial early increases in the ELBO. However, as described in Section \ref{compcons}, there is a trade-off that must be made - models with laps = 1, for example, get close to convergence very fast but then take more time to fully converge. The inclusion of frequent merge/delete moves at the start of the algorithm can also lead to the algorithm jumping to a worse local optimum, reflected in slightly lower ARIs in some cases.
   
\textbf{Better ARIs are achieved with merge/delete: } Figure \ref{merdelARIplot} and Appendix \ref{suppfreqresults} show that laps = {2, 5, 10} generally yield higher ARIs, particularly in Simulations 1.1-1.3. 

\begin{figure}[ht!]
\floatconts
  {merdelARIplot}  
  {\caption{Scatter plot comparing ARIs achieved by each model across all datasets and initialisations in Simulations 1.1, 1.2 and 1.3 (labelled 1, 2, 3 respectively). Each point represents one ARI from one MerDel run.}}
  {\includegraphics[width=0.7\linewidth]{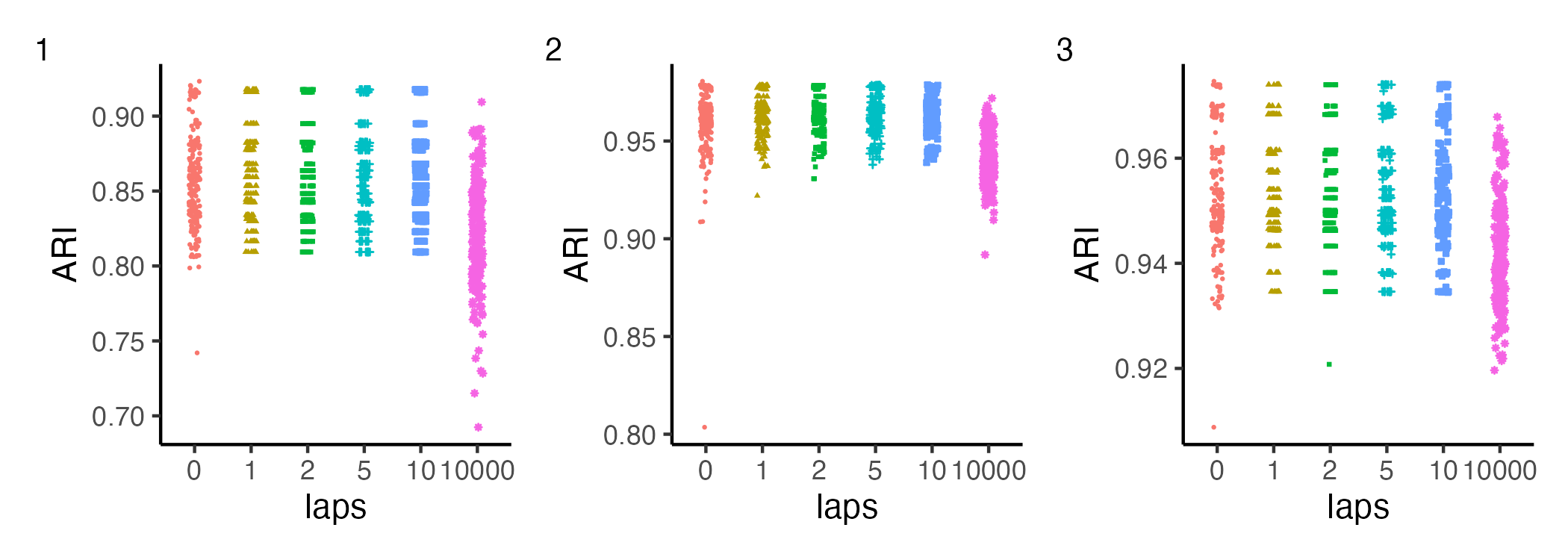}}
\end{figure}

    \textbf{Merge/delete more accurately estimates the number of clusters: } The model with no merge/delete moves tends to vastly overestimate the number of true clusters in the data due to the existence of small unnecessary clusters. MerDel mitigates this, with Figure \ref{merdelsimclustplots} in Appendix \ref{suppfreqresults} showing that the model often is able to find the exact true number. 
    
    \textbf{Variational moves are still necessary: } The algorithm with purely merge/delete moves performs less well; any result of this algorithm is extremely dependent on initialisation. Observations can never move between two clusters. 

Table \ref{tablesim_combined} shows wall-clock time, final log-ELBO and ARI of the resulting clustering structures in Simulations 1.1, 1.2 and 1.3.

\begin{table*}[htb!]
\floatconts
  {tablesim_combined}
  {\caption{Performance comparison across Simulations 1.1, 1.2, and 1.3: mean wall-clock time, log-ELBO, ARI and number of clusters with confidence intervals for time and ARI (±1.96 × standard deviation).}}
  {\footnotesize
  \renewcommand{\arraystretch}{0.9}
  \begin{tabular}{llcccc}
\toprule
\bfseries Simulation & \bfseries Laps & \bfseries Time (s) & \bfseries log-ELBO & \bfseries ARI & \bfseries Clusters \\
\midrule
1.1 & 0 & \textbf{3.46 [3.43, 3.49]} & -25357 & 0.856 [0.852, 0.860] & 5.01\\
& 1 & 6.35 [6.16, 6.55] & \textbf{-25354} & \textbf{0.858 [0.854, 0.862]} & 5.00 \\
& 2 & 5.88 [5.75, 6.01] & \textbf{-25354} & \textbf{0.858 [0.854, 0.862]} & 5.00\\
& 5 & 6.75 [6.65, 6.85] & \textbf{-25354} & \textbf{0.858 [0.854, 0.862]} & 5.00\\
& 10 & 8.68 [8.57, 8.80] & \textbf{-25354} & \textbf{0.858 [0.854, 0.862]} & 5.00 \\
& 10000 & 14.0 [13.2, 14.7] & -25617 & 0.823 [0.818, 0.828] & 15.3\\
\midrule
1.2 & 0 & \textbf{10.9 [10.8, 11.0]} & -81256 & 0.959 [0.957, 0.961] & 7.65\\
& 1 & 29.1 [26.4, 31.8] & -81245 & 0.962 [0.960, 0.963] & 7.63\\
& 2 & 23.4 [21.7, 25.1] & -81238 & 0.962 [0.961, 0.963] & 7.70\\
& 5 & 25.4 [24.1, 26.7] & \textbf{-81236} & \textbf{0.963 [0.962, 0.964]} & 7.84\\
& 10 & 31.0 [30.0, 32.0] & -81260 & 0.962 [0.961, 0.963] & 8.54\\
& 10000 & 47.5 [45.0, 50.0] & -81721 & 0.940 [0.938, 0.942] & 18.0\\
\midrule
1.3 & 0 & \textbf{29.1 [28.8, 29.4]} & -164099 & 0.953 [0.952, 0.955] & 10.0\\
& 1 & 57.1 [54.8, 59.5] & \textbf{-164090} & \textbf{0.954 [0.953, 0.955]} & 10.0\\
& 2 & 52.1 [50.7, 53.5] & -164095 & 0.954 [0.952, 0.955] & 10.0\\
& 5 & 61.8 [60.5, 63.1] & -164100 & 0.954 [0.952, 0.955] & 10.2\\
& 10 & 76.5 [74.7, 78.3] & -164173 & 0.953 [0.951, 0.954] & 12.0\\
& 10000 & 122 [112, 113] & -164675 & 0.943 [0.941, 0.944] & 22.7\\
\bottomrule
  \end{tabular}}
\end{table*}

\begin{figure}[ht]
\centerline{\includegraphics[scale=0.6]{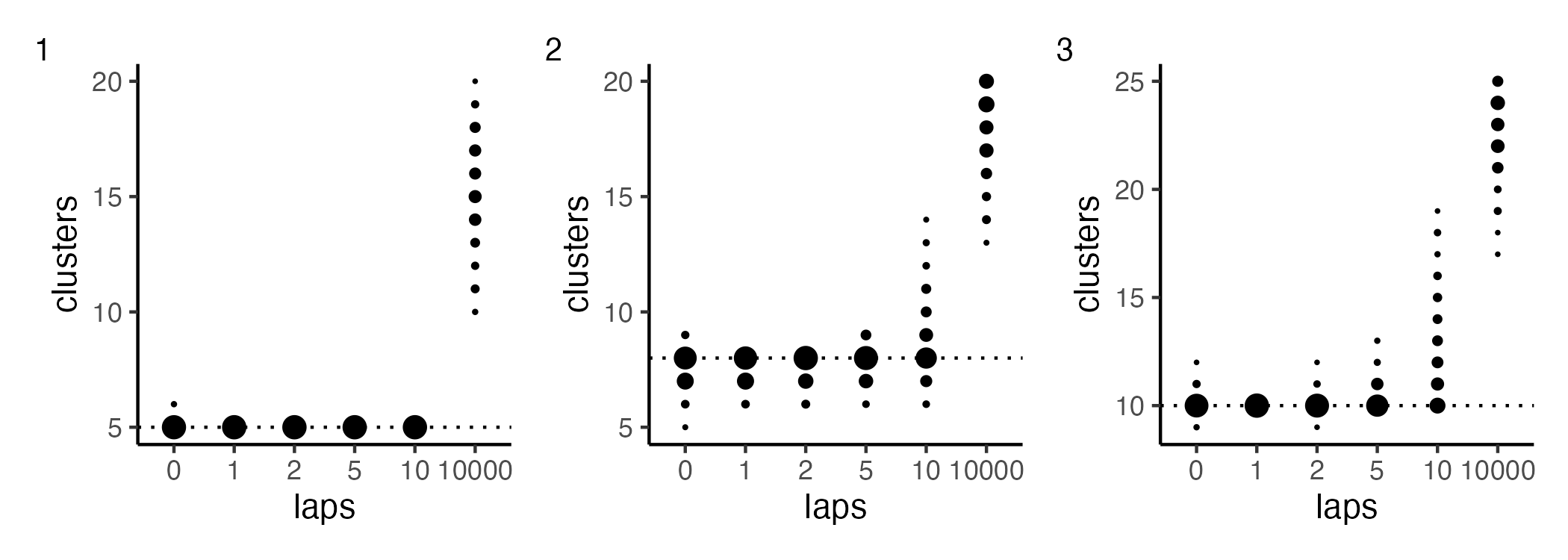}}
\caption{Plot comparing final number of clusters in clustering models in simulations 1.1-1.3.} \label{merdelsimclustplots}
\end{figure}

\begin{figure}[!ht]
    \centering

    \begin{minipage}[b]{0.44\textwidth}
        \centering
        \includegraphics[width=\linewidth]{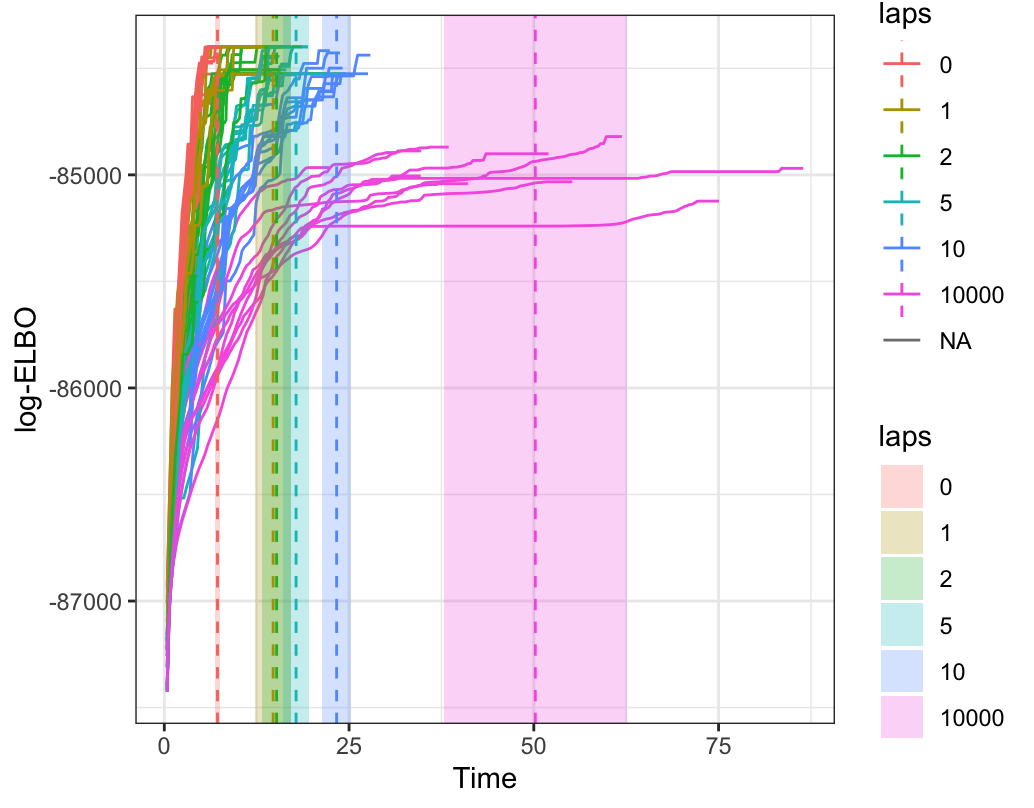}
        \caption*{(a) Simulation 1.2}
    \end{minipage}
    \hfill
    \begin{minipage}[b]{0.44\textwidth}
        \centering
        \includegraphics[width=\linewidth]{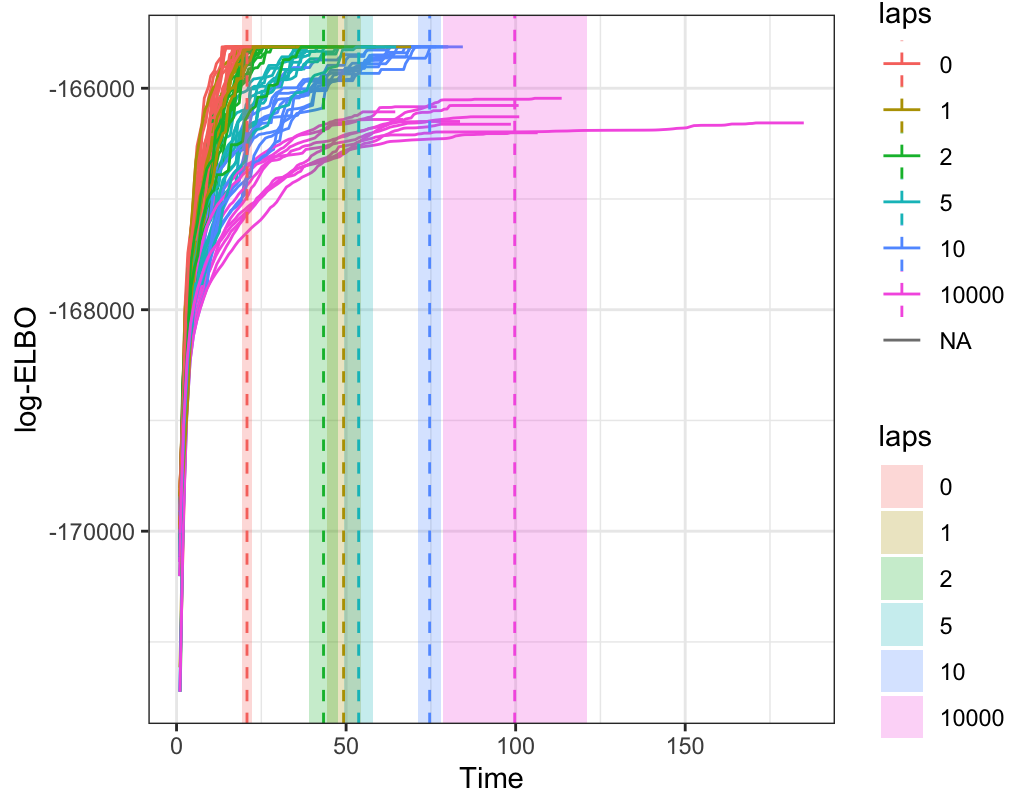}
        \caption*{(b) Simulation 1.3}
    \end{minipage}

    \caption{Graphs showing the ELBO vs wall-clock time for the algorithm for two simulated datasets across all 20 initialisations for each of the different "laps" in Simulations 1.2 and 1.3. The mean end time across all initialisations is shown, as well as an approximate 95\% confidence interval.}
    \label{Sim1linegraphs}
\end{figure}
We can clearly see from Figure \ref{Sim1linegraphs} that a) the original variational model clearly takes far longer than models with merge/delete b) there are steep improvements in ELBO near the start with more frequent merge/delete moves. However, for example, `laps = 1' is near convergence very early on in both plots, but the algorithm takes longer to actually converge and stop. 

Another observation we made, especially with Simulation 1.1, is that the variations of MerDel (laps between 1 to 10) frequently identified exactly the same model as one another and achieved higher ARIs than the variational models with no merge/delete moves. This could be indicative that the merge/delete moves are allowing the model to escape local optima and could be reaching a global optimum.

\subsubsection{Variable Selection}

\begin{figure}[ht]
\centerline{\includegraphics[scale=0.65]{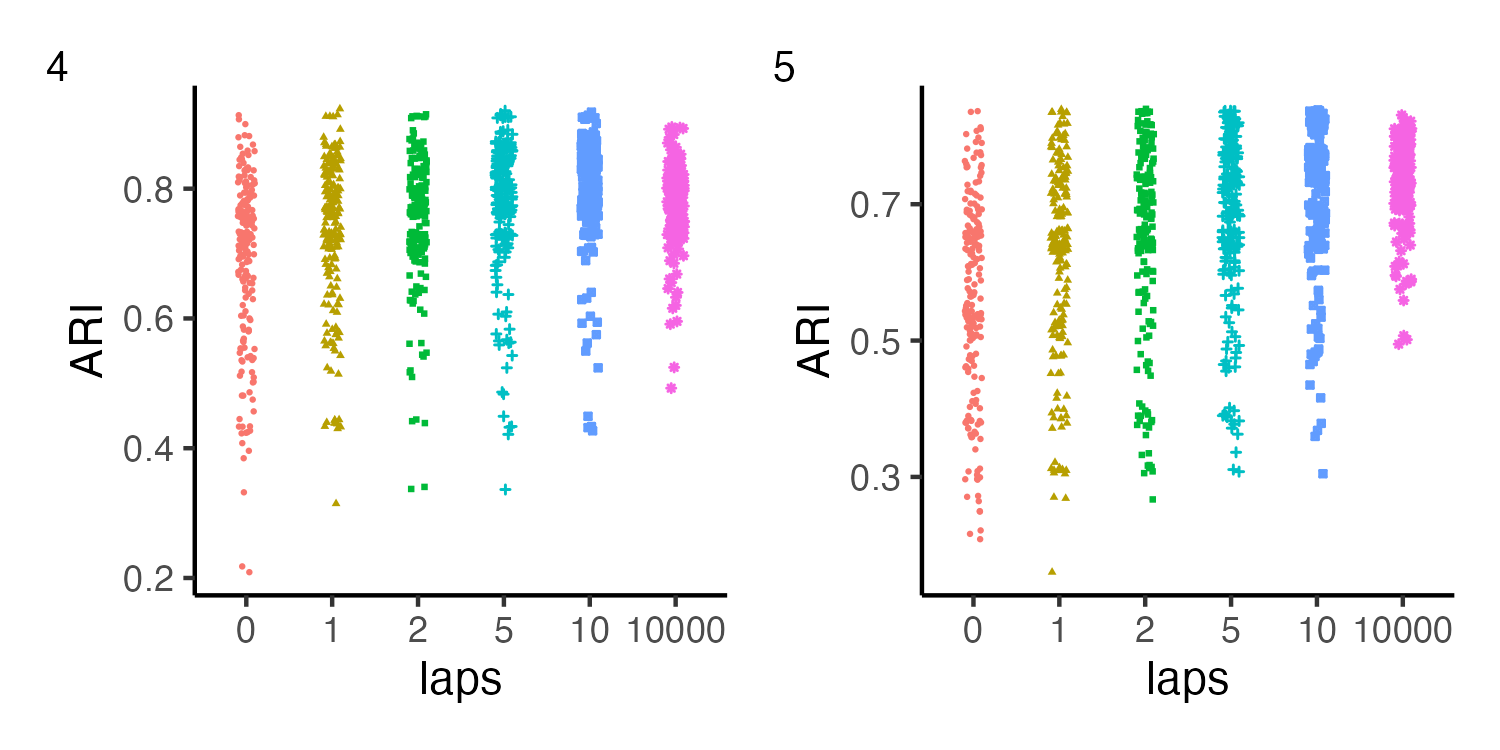}}
\caption{Scatter plot comparing ARIs across all datasets and initialisations in Simulations 1.4 and 1.5.}\label{merdelvsARIplot}
\end{figure}

Table \ref{tablesim45_combined} shows wall-clock time, final log-ELBO and ARI of the resulting clustering structures in Simulations 1.4 and 1.5. We still see a speed up in terms of wall-clock time in these simulations, but see that there is more inconsistency with solutions with noisier data, especially as we increase the number of merge/delete moves, and the mean ARI for merge/delete models is generally lower when we incorporate more merge/delete moves. The number of clusters at convergence is also lower than expected. We see from Figure \ref{merdelvsARIplot} that it seems to be that merge/delete moves lead to less consistent results in the variable selection setting, but most of the best-performing models in terms of ARI are from MerDel models. 


\begin{table*}[htb!]
\floatconts
  {tablesim45_combined}
  {\caption{Performance comparison across Simulations 1.4 and 1.5: mean wall-clock time, log-ELBO, ARI and number of clusters with confidence intervals for time and ARI (±1.96 × standard deviation).}}
  {\footnotesize
  \renewcommand{\arraystretch}{0.9}
  \begin{tabular}{llcccc}
\toprule
\bfseries Simulation & \bfseries Laps & \bfseries Time (s) & \bfseries log-ELBO & \bfseries ARI & \bfseries Clusters \\
\midrule
1.4 & 0 & \textbf{16.9 [16.6, 17.2]} & -41913 & 0.692 [0.674, 0.710] & 7.53\\
& 1 & 91.7 [84.0, 99.3] & -41848 & 0.731 [0.714, 0.747] & 7.86\\
& 2 & 75.6 [69.8, 81.4] & -41838 & 0.753 [0.739, 0.767] & 8.10\\
& 5 & 74.4 [70.0, 78.7] & \textbf{-41823} & 0.773 [0.759, 0.787] & 8.48\\
& 10 & 92.7 [88.4, 96.9] & -41831 & \textbf{0.792 [0.780, 0.804]} & 9.16\\
& 10000 & 128 [118, 138] & -42372 & 0.782 [0.774, 0.791] & 19.3\\
\midrule
1.5 & 0 & \textbf{24.0 [23.7, 24.4]} & -61781 & 0.561 [0.540, 0.581] & 6.60 \\
& 1 & 137 [125, 149] & -61664 & 0.625 [0.605, 0.645] & 7.29 \\
& 2 & 110 [101, 118] & -61640 & 0.651 [0.631, 0.670] & 7.67 \\
& 5 & 112 [107, 118] & -61622 & 0.679 [0.661, 0.696] & 8.12 \\
& 10 & 149 [143, 156] & \textbf{-61612} & 0.710 [0.695, 0.725] & 8.90 \\
& 10000 & 213 [199, 227] & -62204 & \textbf{0.730 [0.721, 0.739]} & 19.3 \\
\bottomrule
  \end{tabular}}
\end{table*}

We additionally used $F_1$ scores to compare the quality of variable selection. Having merge/delete moves tends to improve the quality of the variable selection as quantified by $F_1$ scores. This is illustrated in Table \ref{selectionsim1vs}. The gains are made especially in terms of identifying irrelevant, noisy variables, and laps 2 and 5 get the top $F_1$ scores of 0.977, indicating extremely good feature selection performance.

\begin{table*}[ht]
\caption{Table comparing the mean $F_1$ scores, and number of relevant and irrelevant variables successfully found across all simulated datasets and all initialisations in Simulation 1.4 and Simulation 1.5.}
\label{selectionsim1vs}
\begin{center}
\begin{footnotesize}
\begin{sc}
\begin{tabular}{lccc|ccr}
\toprule
 & \multicolumn{3}{c }{Simulation 1.4} & \multicolumn{3}{|c}{Simulation 1.5} \\
Laps & Rel & Irrel & $F_1$ Score & Rel & Irrel & $F_1$ Score \\
\midrule
0 & 73.4 & 21.7 & 0.968 & 58.8 & 34.6 & 0.947\\
1 & 73.5 & \textbf{22.7} & 0.975 & 58.7 & \textbf{36.0} & 0.957\\
2 & 73.9 & \textbf{22.7} & \textbf{0.977} & 59.2 & 35.8 & \textbf{0.959}\\
5 & 74.0 & 22.4 & \textbf{0.977} & \textbf{59.3} & 35.5 & 0.958\\
10 & 74.0 & 22.2 & 0.975 & \textbf{59.3} & 34.9 & 0.954\\
10000 & \textbf{74.1} & 15.5 & 0.935 & \textbf{59.3} & 23.1 & 0.872\\
\bottomrule
\end{tabular}
\end{sc}
\end{footnotesize}
\end{center}
\end{table*}

\clearpage 
\subsection{Global Merge Simulations - Additional Results} \label{globalmergeappendix}

\begin{table*}[htbp]
\floatconts
  {globalmergeresultsfull}
  {\caption{`Global Merge Simulations' results. We report the median and lower/upper quantiles across all 10 `shuffles'/initialisations and all 10 synthetic datasets. FedMerDel/G is the greedy search, FedMerDel/R is the random search for FedMerDel. Note separate independent datasets were generated for $N$ equal but different numbers of batches/cores, accounting for slight differences in ARI. Heterogeneous simulations (a), (b), (c) are described in Appendix G.4.}}
  {\footnotesize
  \renewcommand{\arraystretch}{0.9}
  \begin{tabular}{lccccc}
\toprule
$\mathbf{N}$ \bfseries/Sim & \bfseries Model & \bfseries Batches & \bfseries ARI & \bfseries Clusters & \bfseries Time (s)\\
\midrule
20000 & full & 5 & 0.943 [0.933, 0.950] & 12 [11.8, 12] & 106 [92.8, 130]\\ 
 & par & 5 & 0.943 [0.930, 0.950] & 12 [11, 12] & 37.0 [33.4, 40.6] \\
 & FedMerDel/G& 5 & 0.920 [0.912, 0.933] & 12 [12, 12] & 33.1 [28.9, 37.2]\\
 & FedMerDel/R & 5 & 0.920 [0.912, 0.932] & 12 [12, 12] & 33.2 [29.1, 37.4]\\
\midrule
50000 & full & 5 & 0.947 [0.941, 0.954] & 12 [12, 12] & 467 [414, 554]\\ 
 & par & 5 & 0.948 [0.942, 0.954] & 12 [12, 12] & 176 [163, 196] \\
 & FedMerDel/G& 5 & 0.945 [0.939, 0.949] & 13 [13, 14] & 114 [106, 123]\\
 & FedMerDel/R & 5 & 0.945 [0.939, 0.948] & 12 [12, 12] & 114 [106, 123]\\
\midrule
50000 & full & 10 & 0.957 [0.955, 0.958] & 12 [12, 12] & 465 [432, 512]\\ 
 & par & 10 & 0.956 [0.955, 0.958] & 12 [12, 12] & 126 [121, 142] \\
 & FedMerDel/G& 10 & 0.951 [0.946, 0.954] & 13 [13, 14] & 70.6 [66.2, 75.8]\\
 & FedMerDel/R & 10 & 0.951 [0.946, 0.954] & 12 [12, 13] & 71.3 [66.8, 76.6]\\
\midrule
100000 & full & 5 & 0.955 [0.944, 0.961] & 12 [12, 12] & 956 [861, 1116]\\
 & par & 5 & 0.954 [0.944, 0.961] & 12 [12, 12] & 338 [310, 380] \\
 & FedMerDel/G& 5 & 0.954 [0.943, 0.960] & 13 [12, 13] & 260 [231, 287]\\
 & FedMerDel/R & 5 & 0.954 [0.943, 0.960] & 12 [12, 12] & 260 [232, 287]\\
\midrule
100000 & par & 10 & 0.949 [0.941, 0.959] & 12 [12, 12] & 249 [231, 284] \\
 & FedMerDel/G& 10 & 0.948 [0.938, 0.952] & 13 [13, 14] & 131 [123, 143]\\
 & FedMerDel/R & 10 & 0.948 [0.938, 0.952] & 12 [12, 13] & 132 [124, 144]\\
\midrule
200000 & par & 10 & 0.952 [0.947, 0.958] & 12 [12, 12] & 482 [446, 536]\\
 & FedMerDel/G& 10 & 0.951 [0.946, 0.957] & 13 [13, 14] & 269 [242, 303]\\
& FedMerDel/R & 10 & 0.951 [0.946, 0.957] & 12.5 [12, 13] & 270 [243, 305]\\
\midrule
500000 & par & 10 & 0.950 [0.943, 0.951] & 12 [12, 12] & 1121 [1056, 1303]\\
 & FedMerDel/G& 10 & 0.949 [0.942, 0.951] & 13 [13, 14] & 685 [642, 793]\\
& FedMerDel/R & 10 & 0.949 [0.942, 0.951] & 12.5 [12, 13] & 688 [646, 797]\\
\midrule
1000000 & par & 20 & 0.948 [0.943, 0.953] & 12 [12, 12] & 2119 [2031, 2204]\\
 & FedMerDel/G& 20 & 0.948 [0.943, 0.953] & 13.5 [13, 14] & 899 [825, 984]\\
& FedMerDel/R & 20 & 0.948 [0.943, 0.953] & 14 [13, 14] & 912 [840, 1007]\\
\midrule
(a) & par & 10 & 0.945 [0.944, 0.951] & 12 [12, 12] & 119 [110, 131]\\  
  & FedMerDel & 10 & 0.942 [0.936, 0.946] & 13 [12, 13] & 64.5 [61.5, 70.1] \\
\midrule
(b) & par & 5 & 0.955 [0.949, 0.956] & 10 [10, 10] & 169 [161, 192] \\
 & FedMerDel & 5 & 0.993 [0.992, 0.994] & 10 [10, 10] & 131 [127, 165] \\
\midrule
(c) & par & 5 & 0.950 [0.944, 0.950] & 12 [12, 12.8] & 69.5 [65.8, 77.3] \\
 & FedMerDel & 5 & 0.988 [0.985, 0.990] & 13 [12, 13] & 53.2 [48.3, 58.3] \\
\bottomrule
  \end{tabular}}
\end{table*}

\subsubsection{Comparisons to Other Unsupervised Centralised Learning Methods} \label{appendix:unsupervisedcomp}

\textcolor{black}{While k-modes, hierarchical clustering and DBSCAN were fast and clearly superior in terms of computational efficiency (eg. k-modes took around 25 seconds on datasets of size $N=50,000$, all performed very poorly with ARIs close to 0 compared to the true simulated labels, suggesting these were barely better than random cluster assignment. These methods also face limitations; they often require manual specification of the number of clusters, have limited interpretability and struggle with high dimensionality.}

\textcolor{black}{Using the poLCA R package on a simulated dataset with N = 50,000, LCA achieved a mean ARI of 0.867, underperforming both FedMerDel and parallelised MerDel. Furthermore, this required around 10 minutes per model. LCA requires users to fit multiple models and compare with criteria such as BIC to determine the number of clusters. Crucially, this method is not directly applicable to FL scenarios.}




\subsubsection{Varying P} \label{globalmergeresultsP}

We saw a similar pattern to $N$ where as $P$ increases, the gains made from FedMerDel are increased compared to parallelised MerDel.

\begin{table*}[htbp]
\floatconts
  {globalmergeresultsP}
  {\caption{`Global Merge Simulations' results, where all datasets are of size $N = 100,000$ (split into 5 batches/parallelised over 5 cores) and have 10 equally sized clusters, but $P$ (number of covariates) is varied. We compare FedMerDel to parallelised MerDel. We report the median and lower/upper quantiles across 5 `shuffles'/initialisations for 5 synthetic datasets. Results are for the random search; greedy searches differed insignificantly.}}
  {\footnotesize
  \renewcommand{\arraystretch}{0.9}
  \begin{tabular}{lcccc}
\toprule
$\mathbf{N}$ & \bfseries Model & \bfseries ARI & \bfseries Clusters & \bfseries Time (s) \\
\midrule
60 & par & 0.782 [0.753, 0.796] & 10 [10, 10] & 206 [194, 231]\\
 & FedMerDel & 0.780 [0.751, 0.794] & 10 [10, 10] & 153 [130, 169] \\
\midrule
80 & par & 0.874 [0.872, 0.880] & 10 [10, 10] & 285 [256, 333]\\
 & FedMerDel & 0.872 [0.870, 0.879] & 10 [10, 10] & 190 [182, 231] \\
\midrule
120 & par & 0.972 [0.972, 0.973] & 10 [10, 10] & 316 [292, 345]\\
 & FedMerDel & 0.972 [0.971, 0.972] & 10 [10, 10] & 208 [203, 229]\\
\midrule
200 & par & 0.999 [0.999, 0.999] & 10 [10, 10] & 465 [453, 469]\\
 & FedMerDel & 0.999 [0.999, 0.999] & 10 [10, 10] & 295 [291, 302]\\
\bottomrule
  \end{tabular}}
\end{table*}

\begin{figure}[ht]
\centerline{\includegraphics[scale=0.45]{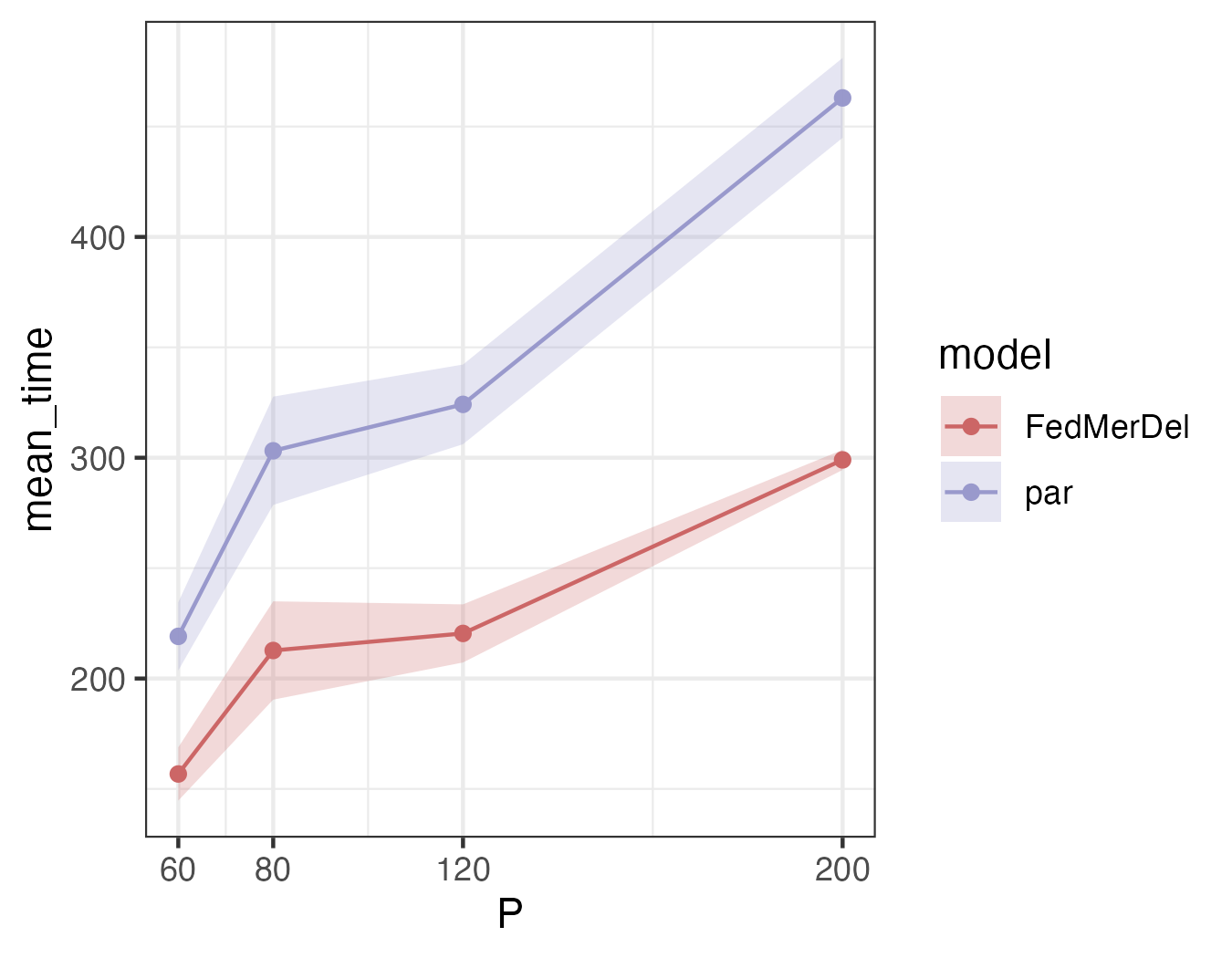}}
\caption{Plot comparing the mean time taken by parallelised MerDel (`par') and FedMerDel (`shard') as we vary $P$, with a 95\% confidence interval (mean $\pm$ 1.96 $\times \frac{s.d}{\sqrt{n}}$). Results for FedMerDel are with the greedy search.}
\end{figure}

\subsubsection{Less Separable Clusters}
\textcolor{black}{In both simulation scenarios with the alternative generating distribution to allow for less cluster separability, results still remained robust as before, where ARIs remained close to 0.99. Furthermore, in the second scenario where 75\% of covariates were similar across clusters 11 and 12, in all cases, FedMerDel separated the observations from these clusters into distinct clusters.}

\subsubsection{Global Merge with Variable Selection}

Results showed similar performance in terms of ARI between FedMerDel and MerDel on the full dataset (Table \ref{globalmergeresultsvs}). In these scenarios (with relatively low $N$) FedMerDel notably was almost always able to correctly identify relevant and irrelevant variables, outperforming other methods, which usually wrongly identified some irrelevant variables as relevant. Global merges took less than one second in all cases.

\begin{table*}[h!]
\floatconts
  {globalmergeresultsvs}
  {\caption{`Global Merge with Variable Selection' simulation results. We report the median and lower/upper quantiles across all 5 `shuffles'/initialisations and all 5 synthetic datasets. Global merge uses the `greedy search'. `Rel' and `Irrel' refer to the number of correctly identified relevant and irrelevant covariates respectively.}}
  {\footnotesize
  \renewcommand{\arraystretch}{0.9}
  \begin{tabular}{lcccccc}
\toprule
$\mathbf{N}$ & \bfseries Model & \bfseries ARI & \bfseries Clusters & \bfseries Time (s) & \bfseries Rel & \bfseries Irrel \\
\midrule
20000 & full & 0.865 [0.862, 0.868] & 15 [15, 15] & 782 [687, 835] & 80 [80, 80] & 18 [17, 18] \\
 & par & 0.865 [0.862, 0.868] & 15 [15, 15] & 207 [187, 228]  & 80 [80, 80] & 17 [16, 17] \\
 & FedMerDel & 0.850 [0.844, 0.857] & 15 [15, 15] & 223 [212, 242]  & 80 [80, 80] & 20 [20, 20] \\
\midrule
50000 & full & 0.885 [0.876, 0.885] & 10 [10, 10] & 1607 [1493, 1767] & 75 [75, 75] & 23 [22, 24] \\
 & par & 0.885 [0.876, 0.886] & 10 [10, 10] & 462 [440, 527]  & 75 [75, 75] & 23 [22, 24] \\
 & FedMerDel & 0.881 [0.873, 0.883] & 10 [10, 10] & 630 [563, 672]  & 75 [75, 75] & 25 [25, 25] \\
\bottomrule
  \end{tabular}}
\end{table*}

\subsection{Number of Batches} \label{result:numberofbatches}

\begin{figure}[htp]
\centering

\begin{minipage}[b]{0.3\textwidth}
    \centering
    \includegraphics[width=\linewidth]{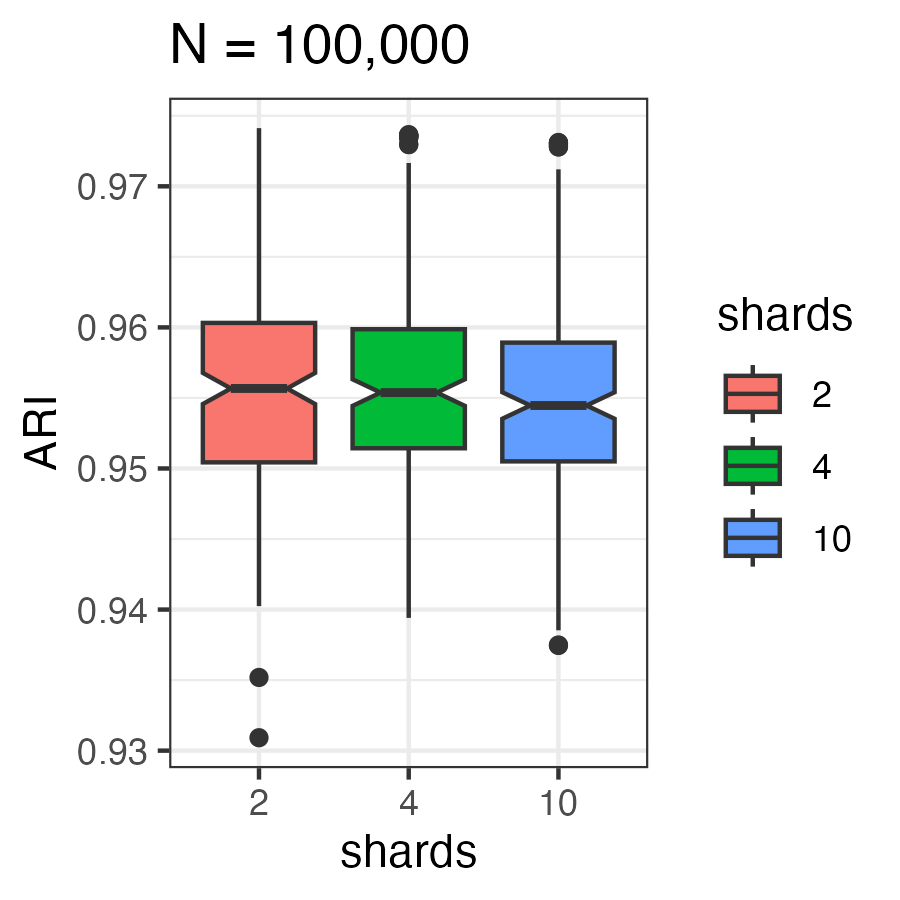}
\end{minipage}
\hspace{1cm}
\begin{minipage}[b]{0.3\textwidth}
    \centering
    \includegraphics[width=\linewidth]{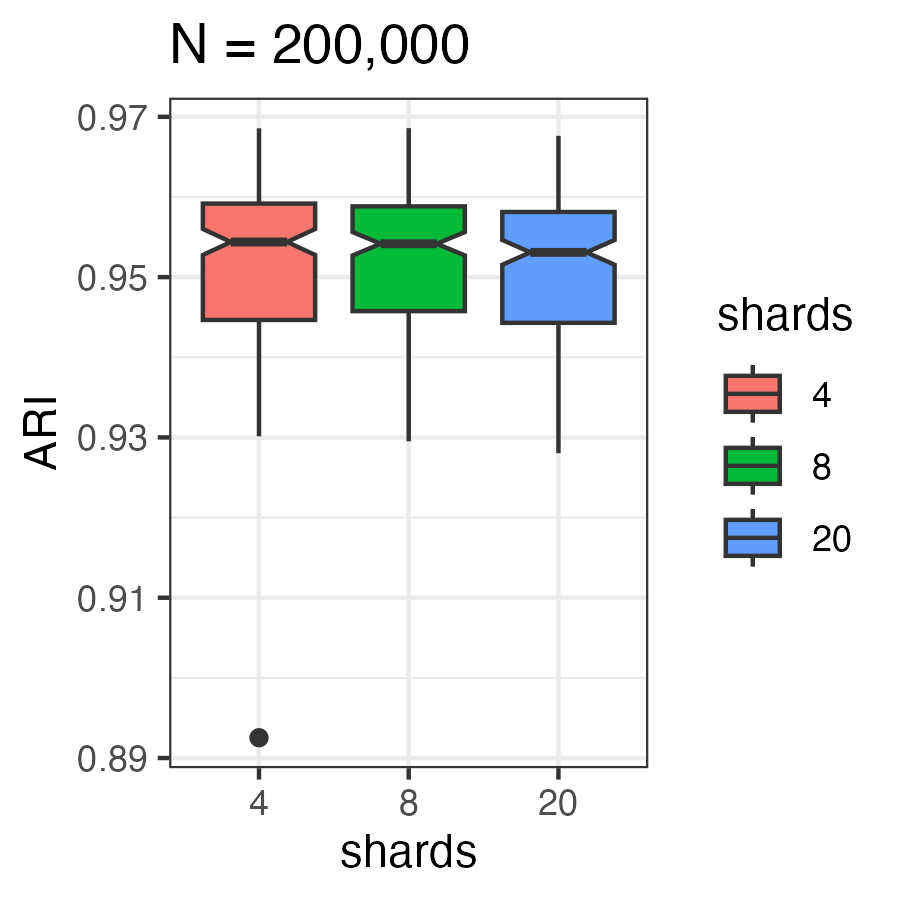}
\end{minipage}

\caption{Boxplot showing the distribution of ARIs across different numbers of batches in the `Number of Batches' simulation.}
\label{noofshardsboxplot}
\end{figure}

Results from this simulation study are visualised in Figure \ref{noofshardsboxplot}. A Kruskal-Wallis test testing for any significant difference between the ARI for different numbers of batches for $N=100,000$ gives a p-value of 0.125. For $N=200,000$, the p-value is 0.0938. One way to potentially improve inference for the number of clusters for higher values of $K_{\mbox{init}}$ would be to look at higher frequencies of merge/delete moves.



\subsection{Comparing Merge Criteria} \label{sec:corrdivrand}

In this simulation, we compared methods for selecting candidate clusters for merging (Section \ref{howtomerge}): correlation (corr'), Kullback-Leibler divergence (KL'), Bhattacharyya distance (bha'), and random selection (rand').

\begin{figure}[htp]
\centering

\begin{minipage}[b]{\textwidth}
  \centering
  \includegraphics[clip,width=0.75\textwidth]{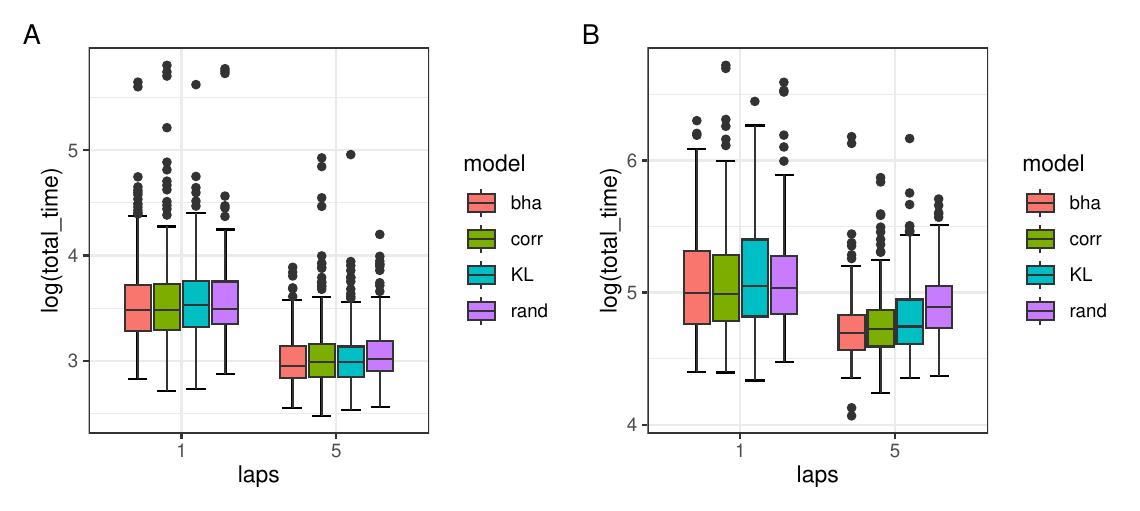}
  \caption*{(a) Comparing log(total time (s))} 
\end{minipage}

\vspace{0.3cm} 

\begin{minipage}[b]{\textwidth}
  \centering
  \includegraphics[clip,width=0.75\textwidth]{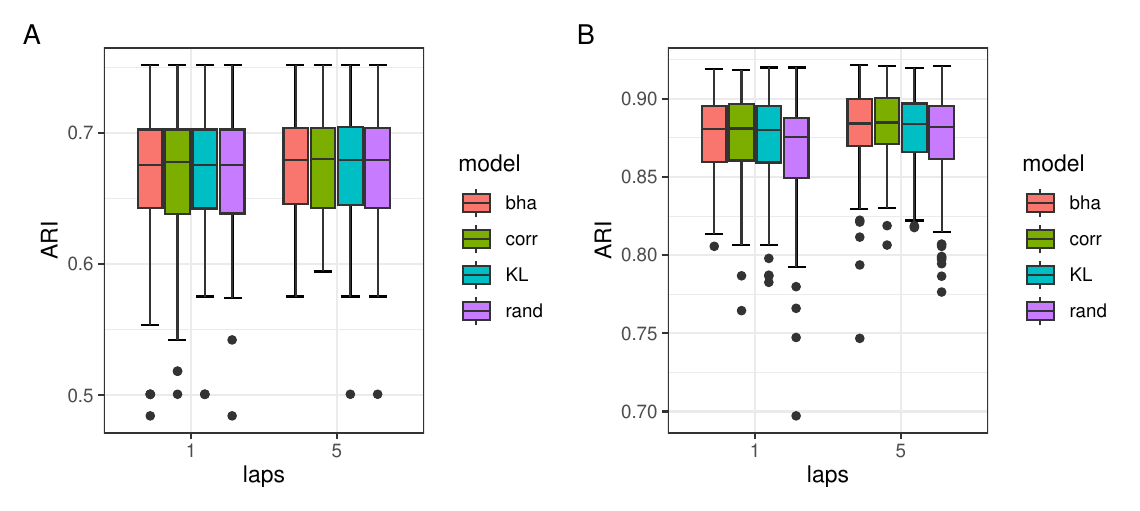}
  \caption*{(b) Comparing ARI} 
\end{minipage}

\vspace{0.3cm} 

\begin{minipage}[b]{\textwidth}
  \centering
  \includegraphics[clip,width=0.75\textwidth]{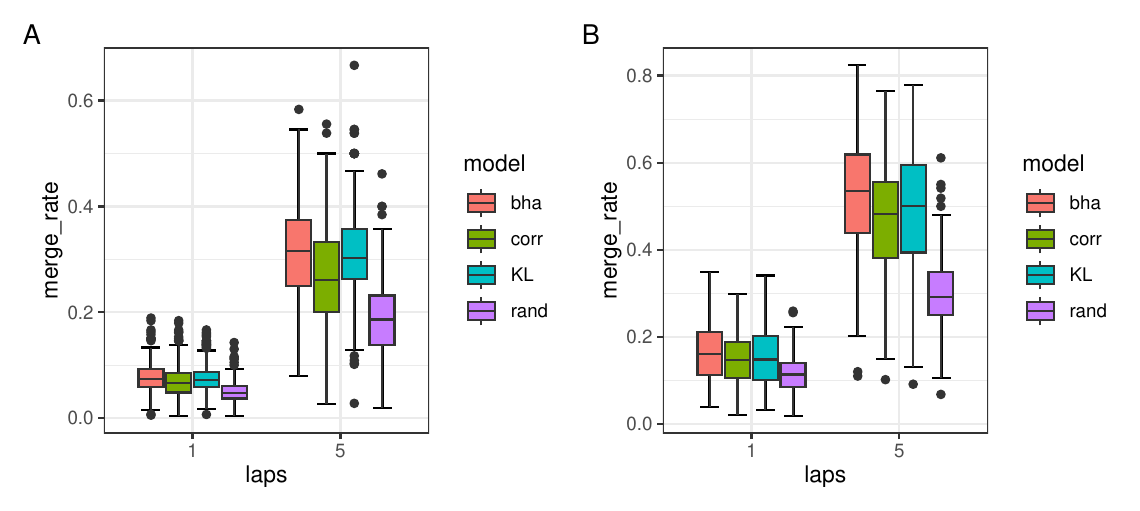}
  \caption*{(c) Comparing `merge rate', defined as proportions of merges accepted.} 
\end{minipage}

\caption{Boxplots comparing the distributions of ARI scores, (the logarithm of) total wall-clock time until convergence, and `merge rates'. All initialisations for each model, simulated dataset, and `laps' value are included, where the boxplots illustrate the median, quantiles, and range across all clustering solutions.}
\label{cdrplots}
\end{figure}

We ran two studies with $N=2000$ and different complexity: (A) $K_{\mbox{true}}=10$, $K_{\mbox{init}}=20$, $P=50$; (B) $K_{\mbox{true}}=20$, $K_{\mbox{init}}=40$, $P=100$ (Figure \ref{cdrplots}).  In both cases, we simulated 30 independently generated binary datasets. For each dataset, we tested 4 models with 10 initializations each and laps = {1, 5}.

Figure \ref{cdrplots} shows no significant differences between methods in time and accuracy, though random selection took slightly longer (especially with laps = 5 and more clusters) and achieved slightly lower median ARIs in study B. Random selection had significantly lower merge acceptance rates, while correlation had lower rates than divergence measures. However, higher computational complexity of divergences resulted in similar total algorithm times.

\clearpage

\subsection{Categorical Data Simulations}

\subsubsection{Frequency of Merge/Delete Moves} \label{sec:catfreq}

Details of the scenarios for this simulation study with categorical data (analogous to Section \ref{setup:freqmerdel}) are given in a table below. Both simulations were run without variable selection. In both scenarios, we generated 20 independent datasets and compared 10 different initialisations. Table~\ref{catsim_combined} shows results; the model with no merge/delete moves (laps = 10000) performed markedly worse in ARI. 

\begin{table*}[htbp]
\floatconts
  {simFedMerDeltablecat}
  {\caption{Parameters for data generation for frequency of merge/delete moves with categorical data.}}
  {\footnotesize
  \renewcommand{\arraystretch}{0.9}
  \begin{tabular}{lcccccr}
\toprule
\bfseries ID & $\mathbf{N}$ & $\mathbf{P}$ & $\mathbf{K_{init}}$ & $\mathbf{K_{true}}$ & $\mathbf{N}$ \bfseries per & \bfseries Categories\\
  &  &  &  &  & \bfseries Cluster & \\
\midrule
    cat.1  & 2000 & 100 & 20 & 8 & 50-800 & 4\\
    cat.2  & 4000 & 100 & 25 & 10 & 200-800 & 3\\
\bottomrule
  \end{tabular}}
\end{table*}





\begin{table*}[htbp]
\floatconts
  {catsim_combined}
  {\caption{Comparisons of mean wall-clock time, mean final log-ELBO, mean ARI and mean number of clusters of final labelling structure across all runs in Simulations cat.1, cat.2 with confidence intervals for time and ARI (+- 1.96 x standard deviation).}}
  {\footnotesize
  \renewcommand{\arraystretch}{0.9}
  \begin{tabular}{llcccc}
\toprule
\bfseries Simulation & \bfseries Laps & \bfseries Time (s) & \bfseries log-ELBO & \bfseries ARI & \bfseries Clusters \\
\midrule
cat.1 & 0 & \textbf{14.4 [14.2, 14.5]} & -238524 & 0.961 [0.951, 0.971] & 7.72\\
& 1 & 27.6 [25.6, 29.5] & -238378 & 0.994 [0.993, 0.995] & 7.48 \\
& 2 & 24.5 [23.1, 25.8] & -238356 & \textbf{0.995 [0.994, 0.996]} & 7.56 \\
& 5 & 29.2 [28.5, 29.8] & \textbf{-238352} & \textbf{0.995 [0.994, 0.996]} & 7.71\\
& 10 & 38.7 [37.8, 39.5] & -238513 & 0.986 [0.984, 0.988] & 8.37\\
& 10000 & 79.2 [74.6, 83.8] & -241501 & 0.831 [0.827, 0.835] & 20\\
\midrule
cat.2 & 0 & \textbf{34.8 [34.1, 35.4]} & -360586 & 0.991 [0.988, 0.995] & 10.1\\
& 1 & 47.0 [46.0, 48.1] & \textbf{-360544} & \textbf{0.999 [0.999, 1.00]} & 10.0\\
& 2 & 49.8 [48.8, 50.7] & \textbf{-360544} & \textbf{0.999 [0.999, 1.00]} & 10.0\\
& 5 & 66.1 [65.1, 67.0] & -360568 & 0.998 [0.998, 0.999] & 10.2\\
& 10 & 91.1 [89.5, 92.8] & -360795 & 0.991 [0.990, 0.992] & 12.1\\
& 10000 & 172 [162, 181] & -312628 & 0.939 [0.937, 0.941] & 25\\
\bottomrule
  \end{tabular}}
\end{table*}

%
%
%
%

\subsubsection{Global Merge Simulations}

We employed random search for global merging using KL divergence to assess cluster similarities. Three scenarios used simulated data with $N =$ 20,000, 50,000, 100,000 observations, $K =$ 20, 10, 12 true clusters, and 4, 5, 5 equally-sized batches respectively. All scenarios had $P = 100$ covariates with 3 categories each. With 5 datasets and 10 runs per model, results showed near-perfect clustering. FedMerDel and parallelised MerDel improved speed, with FedMerDel showing increasing efficiency gains as $N$ increased.

\begin{table*}[htbp]
\floatconts
  {globalmergeresultscat}
  {\caption{`Global Merge Simulations' with categorical data results. We report the median and lower/upper quantiles across all 10 `shuffles'/initialisations and all 5 synthetic datasets. Note the higher merge times and total times with $N=20,000$ is likely due to more initialised clusters (30 vs. 20 in the larger $N$ settings).}}
  {\footnotesize
  \renewcommand{\arraystretch}{0.9}
  \begin{tabular}{lccccc}
\toprule
$\mathbf{N}$ & \bfseries Model & \bfseries ARI & \bfseries Clusters & \bfseries Time (s) & \bfseries Global Merge Time (s)\\
\midrule
20000 & full & 0.999 [0.999, 0.999] & 20 [20, 20] & 387 [373, 418] & \\
& par & 0.999 [0.999, 0.999] & 20 [20, 21] & 146 [128, 157] & \\
 & FedMerDel & 0.998 [0.99, 0.998] & 20 [20, 20] & 124 [118, 131] & \hfil3.91 [3.82, 4.14]\\
\midrule
50000 & full & 0.999 [0.999, 1.00] & 10 [10, 10] & 374 [373, 375]\\
 & par & 0.999 [0.999, 1.00] & 10 [10, 10] & 127 [125, 128] & \\
 & FedMerDel & 0.999 [0.999, 1.00] & 10 [10, 10] & 83.5 [80.7, 84.1] &\hfil 1.29 [1.28, 1.34]\\
 \midrule
100000 & full & 1.00 [0.999, 1.00] & 12 [12, 12] & 946 [857, 1050]\\
 & par & 1.00 [0.999, 1.00] & 12 [12, 12] & 293 [276, 339] & \\
 & FedMerDel & 0.999 [0.990, 1.00] & 12 [12, 13] & 243 [219, 280] &\hfil 2.57 [2.49, 2.64]\\
\bottomrule
  \end{tabular}}
\end{table*}

\clearpage

\section{MNIST Study} \label{appendix:mnist}

\subsection{Simulation Set-Up}
We applied FedMerDel to the MNIST collection of $N=60000$ images of handwritten digits 0-9 \citep{MNIST}. As in \citet{JMLR:v17:11-392}, we downsampled each image to $16\times16$ pixels using bilinear interpolation via TensorFlow, and represented this as a 256-dimensional binary vector. Furthermore, to reduce computation on irrelevant pixels (e.g. those in corners of the image) we removed all pixels with fewer than 100 non-zero values across the whole dataset, leaving 176 covariates . We split this into 6 batches of $N=10000$ and set $K=20$ maximum clusters per batch. True labels for the MNIST digits were discarded.

\subsection{Results}

We ran FedMerDel on 5 different `shuffles' of the data into batches with $K=20$ maximum clusters in each batch, and our resulting models after a global merge consisted of 27-33 clusters. \textcolor{black}{The observed variance in clustering outcomes reflects the genuine difficulty of this unsupervised task - handwritten digits exhibit substantial within-class variation when represented as binary vectors, and this variance is comparable to other unsupervised clustering methods on MNIST. It is unrealistic to expect to find 10 clusters exactly due to the variation of handwritten digits within numbers. The increased number of clusters was not an artefact of the global merging; a run of MerDel on the MNIST test set of $N=10,000$ digits with 35 maximum clusters found between 33-35 clusters where most merges/deletes were rejected. Other unsupervised clustering methods also found a high number of clusters for MNIST \citep{hughes_memoized_2013, Ni2020}.}

Figure \ref{mnistheatmap} depicts the distribution of true number labels within each cluster, and demonstrates subclustering within digits in the `best' model out of our 5 clustering structures. This is chosen as the model with the highest ELBO, an approach used in other variational models \citep{Ueda2002, hughes_memoized_2013}. FedMerDel achieved more consistent performance on certain well-separated digits: for example, 95.95\% of 1's and 92.46\% of 6's were classified into clusters predominantly made up of that digit, while digits with similar shapes (4, 7, 9) were more challenging to distinguish. These numbers have quite similar shapes and have a variety of shapes (Figure \ref{479}). We expected imperfect performance due to the reduction of the image to a 1D vector; every pixel is treated as independent of one another, and we lose all information from the 2D images pertaining to correlation between nearby pixels. Further classification rates can be found in Appendix \ref{sec:classrates}, as well as a comparison to other scalable clustering methods. 

This analysis serves as a supplementary benchmark on real-world data with ground truth labels, while our primary application domain remains EHR data where our simulated experiments demonstrate much lower sensitivity to batch partitioning.


We can use the posterior distribution of our model to probabilistically produce new digits from each cluster, showing the generative potential in variational Bayes; see Figure \ref{fig:gendigits}.

\begin{figure}[htp]
\centering

\begin{minipage}[b]{\textwidth}
  \centering
  \includegraphics[clip,width=0.65\textwidth]{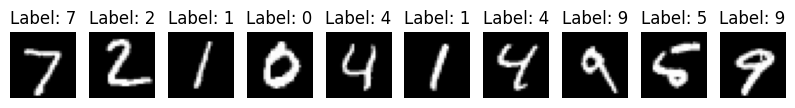}
\end{minipage}
\begin{minipage}[b]{\textwidth}
  \centering
  \includegraphics[clip,width=0.65\textwidth]{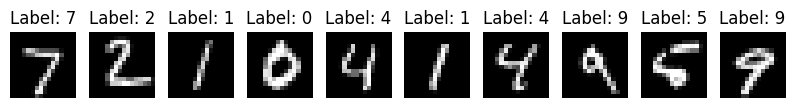}
\end{minipage}
\begin{minipage}[b]{\textwidth}
  \centering
  \includegraphics[clip,width=0.65\textwidth]{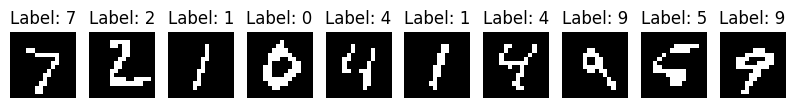}
\end{minipage}

\caption{Example of the first 10 MNIST test-set digits visualised in their original image format (28x28), the same digits with the dimensionality reduction applied (16x16), and the same dimension reduced digits converted to a binary format.}
\end{figure}

\begin{figure}[htp]
\centering

\begin{minipage}[b]{\textwidth}
  \centering
  \includegraphics[clip,width=0.65\textwidth]{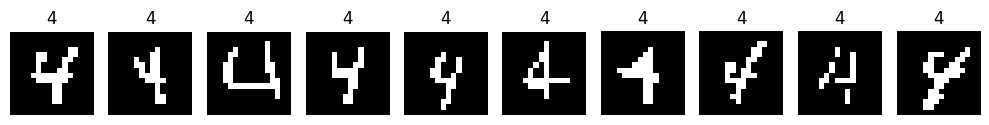}
\end{minipage}

\begin{minipage}[b]{\textwidth}
  \centering
  \includegraphics[clip,width=0.65\textwidth]{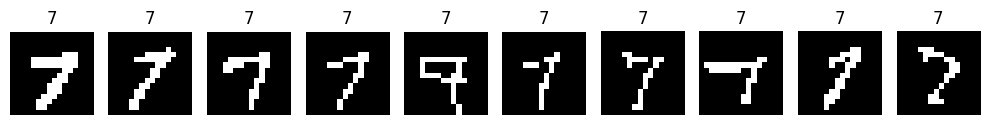}
\end{minipage}

\begin{minipage}[b]{\textwidth}
  \centering
  \includegraphics[clip,width=0.65\textwidth]{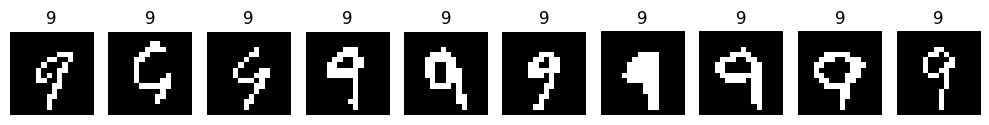}
\end{minipage}
\centering 

\caption{10 randomly selected digits of 4, 7 and 9 from the MNIST dataset, showing similarities in their shapes.} \label{479}

\end{figure}

\begin{figure}[ht]
\centerline{\includegraphics[scale=0.3]{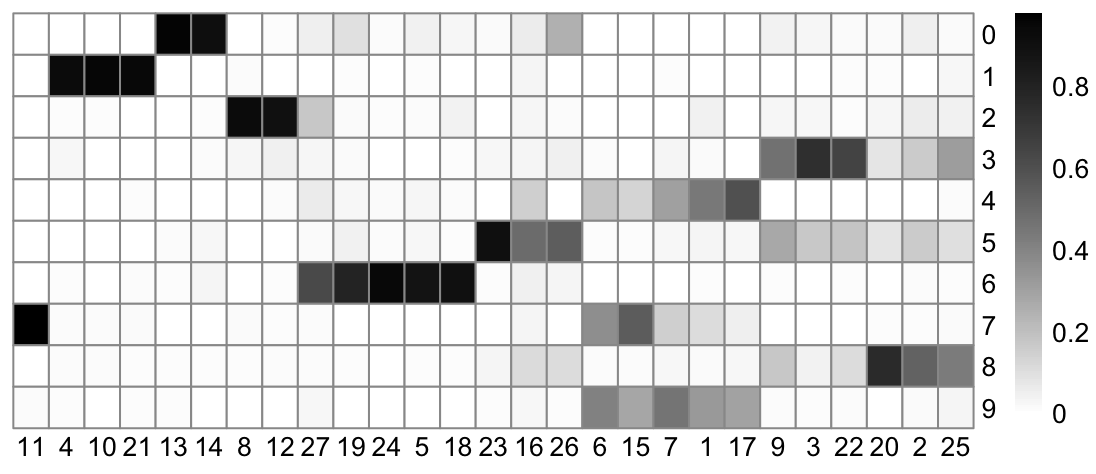}}
\caption{A heatmap showing the correspondence between clusters and true numbers in the clustering model with the best ELBO for the MNIST data. A darker cell colour in entry (i, j) indicates a higher percentage of samples from number i are in the given cluster j. The exact number corresponds to the percentage of cluster j made up of the given number.} \label{mnistheatmap}
\end{figure}

\subsection{Supplementary Results - Classification Rates} \label{sec:classrates}

In Table \ref{mnist_class}, the models included are:
\begin{itemize}
    \item \textbf{FedMerDel\_rand:} FedMerDel, with 6 equally sized batches of data with 20 maximum clusters in each batch. A random search was used for the global merge. All models took between 77-126 mins. 
    \item \textbf{FedMerDel\_greedy:} FedMerDel, but with a greedy search used for the global merge. The same batches were used as with FedMerDel\_rand above. All models took between 77-127 minutes.
    \item \textbf{FedMerDel\_30r}: FedMerDel with 6 equally sized batches of data, but now with 30 maximum clusters in each batch. These models came with high computational time with 60 extra clusters being processed, and reduced scope for interpretation of clusters. Resulting models had between 52-70 clusters. This model used a random search.
    \item \textbf{FedMerDel\_30g}: Same model and batches as above, but with a greedy search. These models had between 52-61 clusters. 
    \item \textbf{kmeans\_xx\_yy}: Data was projected to a lower dimension with Multiple Correspondence Analysis (MCA). We then implemented the k-means algorithm on this lower-dimensional data. xx represents the number of MCA dimensions kept in the model: 10 (explaining 30\% of variance) or 17 (explaining 40\% of variance). yy represents the number for $k$, the number of clusters, where we test 20 and 35. All cases use 20 maximum iterations for k-means, and 20 different initialisations, and present the clustering with the best total within-cluster sum of squares. This method was quick (taking no longer than a minute) and generally performed well, but did not separate numbers such as 2, 3 and 6 as well as FedMerDel.
    \item \textbf{EM\_yy}: We applied an EM algorithm for frequentist model estimation to fit a finite mixture model to the MNIST binary data via the R package \textit{flexmix} \citep{Leisch2004} and used the BIC (Bayesian Information Criterion) for model selection across 10 initialisations. yy represents the number of clusters in the finite mixture model, and we test 20 (10.2 mins) and 35 (55.4 mins).
\end{itemize}
Generally, FedMerDel performed similarly well to comparator methods.

\begin{table}[ht]
\caption{Table comparing classification rates (\%) for each digit across different clustering models. The model analysed further was the 4th FedMerDel\_rand run as the model with the highest ELBO of all the FedMerDel\_rand and FedMerDel\_greedy runs. Classification rate is defined as the \% of digits classified into a cluster primarily consisting of that digit.} \label{mnist_class}
\begin{center}
\begin{footnotesize}
\resizebox{\textwidth}{!}{\begin{tabular}{llllllllllll}
\textbf{MODEL} & \textbf{RUN} & \textbf{0} & \textbf{1} & \textbf{2} & \textbf{3} & \textbf{4} & \textbf{5} & \textbf{6} & \textbf{7} & \textbf{8} & \textbf{9} \\
\hline \\
FedMerDel\_rand & 1 & 83.4 & 95.5 & 83.7 & 75.5 & 49.6 & 37.8 & 90.4 & 62.2 & 65.6 & \textbf{55.3} \\
FedMerDel\_rand & 2 & 84.1 & 94.8 & \textbf{86.3} & 65.2 & 54.8 & 38.6 & 90.1 & 56.0 & 67.3 & 47.8 \\
FedMerDel\_rand & 3 & \textbf{86.4} & 95.4 & 85.8 & 73.9 & 47.6 & 34.6 & \textbf{92.7} & 63.0 & 64.4 & 49.4 \\
FedMerDel\_rand & 4 & 81.4 & \textbf{96.0} & 83.8 & 71.7 & 51.7 & \textbf{47.2} & 92.5 & 67.7 & \textbf{70.2} & 44.8\\
FedMerDel\_rand & 5 & 82.7 & 94.6 & 83.4 & \textbf{83.4} & \textbf{57.7} & 34.5 & 92.0 & \textbf{74.5} & 52.4 & 34.7 \\
\hline
FedMerDel\_greedy & 1 & 83.4 & 95.5 & 83.7 & 75.5 & 53.7 & 37.8 & 90.4 & 62.2 & 65.6 & \textbf{51.4} \\
FedMerDel\_greedy & 2 & 84.1 & 94.8 & \textbf{86.3} & 70.2 & 54.8 & 33.6 & 90.1 & 56.0 & 64.5 & 47.8 \\
FedMerDel\_greedy & 3 & \textbf{86.4} & 95.4 & 85.8 & 73.9 & 47.6 & 34.6 & \textbf{92.7} & 63.0 & 64.4 & 49.4 \\
FedMerDel\_greedy & 4 & 81.4 & \textbf{96.0} & 83.8 & 71.7 & 51.7 & \textbf{47.2} & 92.5 & 67.7 & \textbf{70.2} & 44.8 \\
FedMerDel\_greedy & 5 & 82.7 & 94.6 & 83.4 & \textbf{83.4} & \textbf{57.7} & 34.5 & 92.0 & \textbf{74.5} & 52.4 & 34.6 \\
\hline
FedMerDel\_30r & 1 & \textbf{91.8} & 94.5 & 85.7 & 80.5 & 54.5 & 51.7 & \textbf{94.3} & 71.0 & 68.8 & 50.3 \\
FedMerDel\_30r & 2 & 88.6 & 95.3 & \textbf{85.8} & 76.3 & 56.3 & 46.2 & 93.2 & 70.1 & \textbf{69.0} & \textbf{59.7} \\
FedMerDel\_30r & 3 & 91.3 & \textbf{95.4} & 85.2 & \textbf{83.0} & \textbf{59.8} & \textbf{53.8} & 93.1 & \textbf{78.5} & 65.2 & 34.5 \\
\hline
FedMerDel\_30g & 1 & \textbf{91.8} & 94.5 & 85.7 & 80.5 & 54.5 & 51.7 & \textbf{94.3} & 69.0 & 68.8 & 52.9 \\
FedMerDel\_30g & 2 & 88.6 & 95.3 & \textbf{85.8} & 74.9 & 56.3 & 44.8 & 93.2 & 70.1 & \textbf{71.6} & \textbf{59.7} \\
FedMerDel\_30g & 3 & 91.3 & \textbf{95.4} & 85.2 & \textbf{84.2} & \textbf{62.4} & \textbf{53.8} & 93.1 & \textbf{76.9} & 60.7 & 34.5 \\
\hline 
k-means\_10\_20 & & 81.4 & 97.8 & 65.9 & 74.6 & 54.3 & 31.1 & 83.2 & 70.8 & 40.2 & 39.9 \\
k-means\_10\_35 & & 86.5 & 95.8 & \textbf{73.5} & 72.9 & \textbf{77.6} & \textbf{43.4} & \textbf{86.7} & \textbf{81.2} & 49.5 & 27.7 \\
k-means\_17\_20 & & 83.1 & \textbf{98.6} & 63.7 & 60.9 & 51.9 & 33.4 & 85.0 & 79.7 & 46.3 & 32.6 \\
k-means\_17\_35 & & \textbf{91.5} & 97.8 & 68.3 & \textbf{77.9} & 62.1 & 23.5 & 83.6 & 75.7 & \textbf{57.0} & \textbf{50.0} \\
\hline
EM\_20 & & 82.1 & 95.6 & 83.7 & 84.9 & \textbf{62.2} & 29.7 & 91.8 & 72.4 & 57.4 & 38.0 \\
EM\_35 & & \textbf{87.6} & \textbf{96.1} & \textbf{90.2} & \textbf{85.8} & 51.8 & \textbf{51.9} & \textbf{94.8} & \textbf{81.8} & \textbf{72.3} & \textbf{62.6} \\
\end{tabular}}
\end{footnotesize}
\end{center}
\end{table}

\clearpage

\begin{figure}[htp!]
\centerline{\includegraphics[scale=0.6]{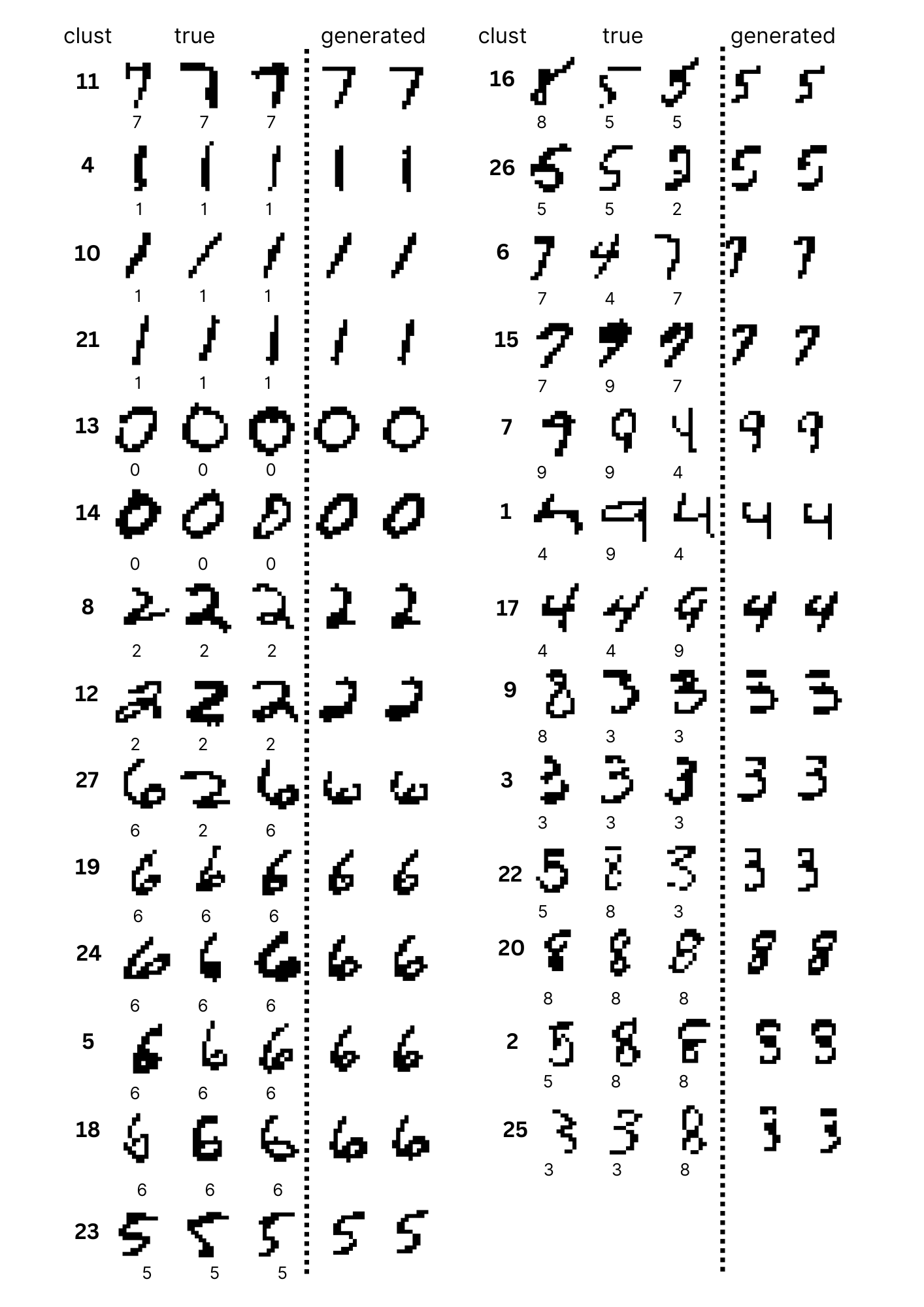}}
\caption{3 randomly selected observations and 2 generated digits (from the variational distribution) from each cluster in the clustering model with the best ELBO for the MNIST data.} \label{fig:gendigits}
\end{figure}

\clearpage

\section{EHR Data - Additional Information} \label{EHRappendix}
EHR data were obtained from The Health Improvement Network database. The start date for our study was 01 January 2010 and end date was 01 January 2017. The dataset comprised primary care health data for 289,821 individuals over 80 years old who had at least two of the long term health conditions considered in our analysis.  The full list of health conditions, and their coding, is provided in Table~\ref{LTCs}. Prevalence of the conditions in Figure \ref{THINsummary} is in Table~\ref{prevalence}. 64.8\% of individuals contributing to the analysis dataset were female, 35.2\% were male.  

\subsection{Characterisation of Random Search Clusters} \label{sec:randomEHRchar}

Note the mean mortality rate across the full dataset was 0.48, and the mean number of conditions per person was 6.19. Mortality rate is defined as the number of people with a registered death date on or before 31/12/2020.

\begin{table*}[htp]
\caption{Characterising the clusters seen in Figure \ref{THINsummary} by the mean number of conditions and mortality rate.}
\label{THINcharacterisation}
\begin{center}
\begin{footnotesize}
\begin{sc}
\begin{tabular}{lcr}
\toprule
Cluster & Mean No. of Conditions & Mortality \\
\midrule
(i) cancer & 6.90 & 0.49 \\
(ii) af \& arrhythmia & 8.48 & 0.57 \\
(iii) pvd \& aortic aneurysm & 8.77 & 0.61 \\
(iv) stroke & 8.46 & 0.54 \\
(v) af \& arrhythmia \& stroke & 11.3 & 0.61 \\
(vi) stroke \& blindness & 12.6 & 0.67 \\
(vii) blindness & 8.69 & 0.53 \\
(viii) dementia & 6.14 & 0.61 \\
(ix) asthma \& copd & 6.30 & 0.54 \\
(x) `generally multimorbid' & 4.57 & 0.39 \\
(xi) `generally multimorbid' & 9.40 & 0.41 \\
(xii) `generally multimorbid' & 7.85 & 0.55 \\
\bottomrule
\end{tabular}
\end{sc}
\end{footnotesize}
\end{center}
\end{table*}

\begin{table}[!ht]
\caption{Table providing prevalence as a percentage for the conditions in Figure \ref{THINsummary}. The most prevalent condition not included in this table is gout, at a prevalence of 7.63\%, which is a condition often found as a comorbidity, but it is not expected that this should form a separate cluster. Other conditions are less common.} \label{prevalence}
    \centering
    \resizebox{0.87\textwidth}{!}{\begin{tabular}{|p{2.8cm}|p{2cm}|p{2.8cm}|p{2cm}|p{2.8cm}|p{2cm}|}
    \hline
    \rowcolor{LightGrey}
        Coding  & Prevalence & Coding  &Prevalence& Coding & Prevalence\\ \hline
        cancerall & 13.8 & tia\_stroke & 18.5 & anxiety & 9.38 \\ \hline
        ca\_skin & 12.6 & eczema & 16.0 & dementia & 13.1 \\ \hline
        arrythmia & 18.3 & deaf & 27.0 & oa & 39.3 \\ \hline
        af & 16.3 & blind & 5.94 & osteoporosis & 14.4  \\ \hline
        hypertension & 63.2 & cataract & 34.9 & ckd & 33.2\\ \hline
        hf & 10.1 & glaucoma & 9.77 &  asthma & 12.2\\ \hline
        ihd & 26.8 & amd & 8.47 & copd & 8.20 \\ \hline
        pvd & 7.64 & retinopathy & 8.08 & hypothyroid & 13.1\\ \hline
        aortic\_aneurysm & 1.84 & diverticuli & 14.6 & t2dm & 14.7 \\ \hline
        tia & 9.55 & depression & 13.6 & bph & 9.43 \\ \hline
        nos\_stroke & 10.4 &&&& \\ \hline 
    \end{tabular}}
\end{table}
\clearpage 
\subsection{Heterogeneous EHR Simulations} \label{sec:hetehr}

\begin{figure}[ht]
\centerline{\includegraphics[scale=0.6]{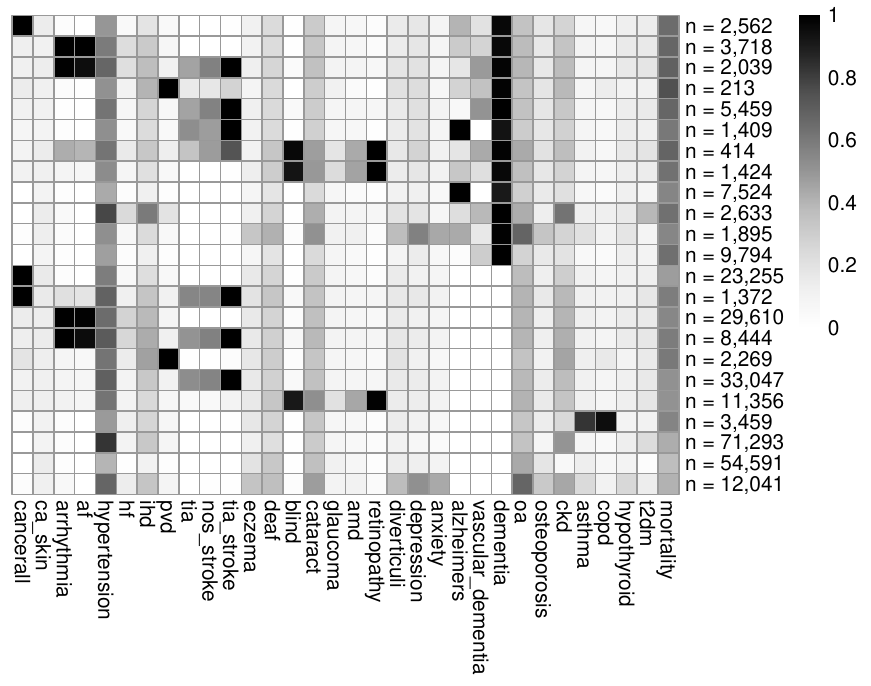}}
\caption{Results of global EHR clustering where individuals with and without dementia were separated into different clusters. 15 total similarly sized batches were randomly generated after the data was divided. Random merge was used. Similar clusters are seen, but split by those with and without dementia as expected. Notably, an asthma/COPD cluster emerges in patients without dementia, but this is not seen for patients with dementia. Similar mortality/mean number of conditions as in Appendix J.1 could be found to infer clinical risk based on having dementia as an additional comorbidity or not. To improve visualisation, only the most prevalent health conditions are shown. }\label{Demsummary1}
\end{figure}

\begin{figure}[ht]
\centerline{\includegraphics[scale=0.6]{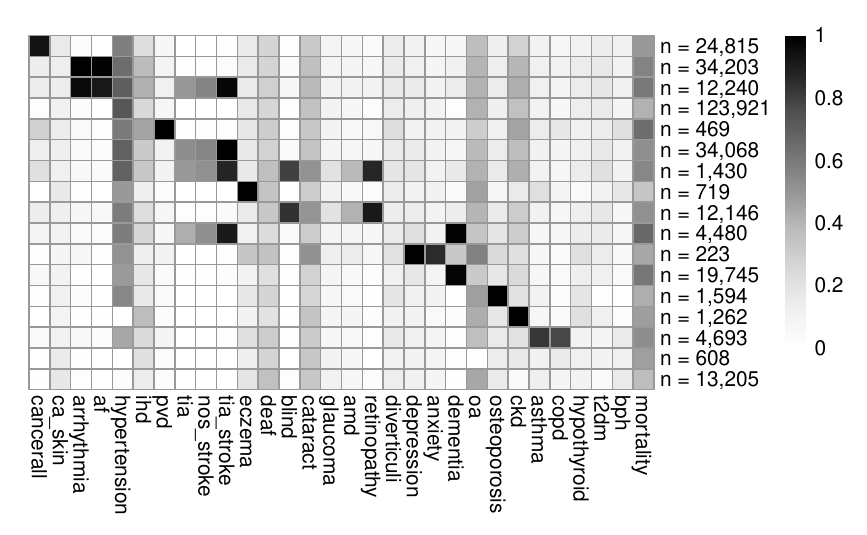}}
\caption{Results of global EHR clustering where a symmetric 15-dimensional Dirichlet(1) distribution was used to simulate heterogeneity. This was done by taking the 12 clusters from the analysis in the main manuscript, and assigning observations from each cluster into 15 new batches according to the Dirichlet distribution. Random merge was used. Core clusters are still clearly retained (eg. cancer, respiratory disorders, stroke only, mixed stroke + AF/arrhythmia) while smaller clusters emerge, likely driven by shard specific overrepresentation. For example, a small multimorbid cluster emerges with no osteoarthritis. To improve visualisation, only the most prevalent health conditions are shown. }\label{Hetsummary2}
\end{figure}

\subsection{Greedy Search Global Merge}
In Figure 3 of the main manuscript, we illustrated the global clusters identified using the ``random search" approach to perform the global merge. In Figure \ref{MMsummary1} below, we provide the corresponding results obtained using the ``greedy search" approach.

\begin{figure}[ht]
\centerline{\includegraphics[scale=0.6]{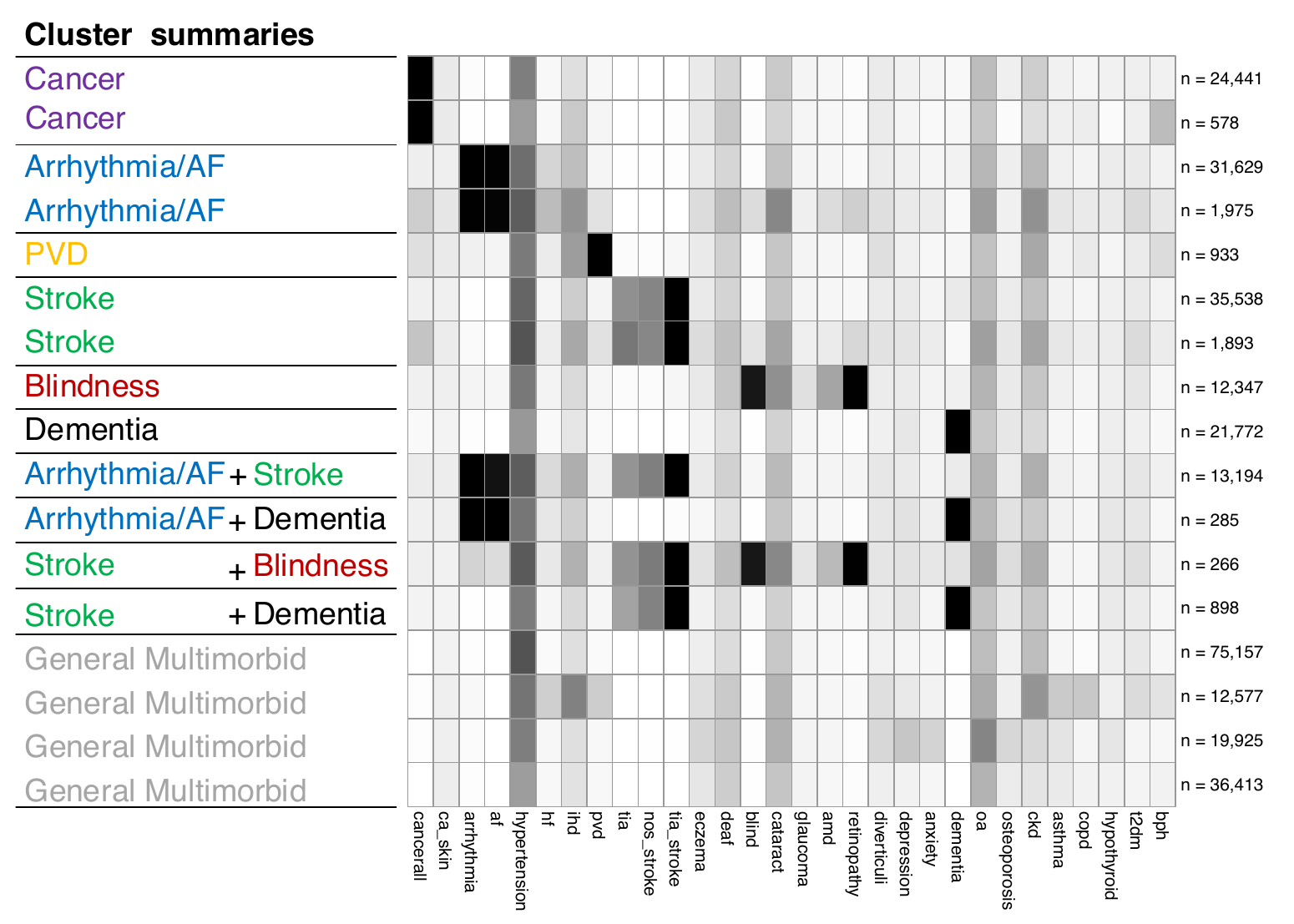}}
\caption{Results of global EHR clustering using the greedy search approach to perform the global merge. Columns are health conditions, and rows correspond to clusters. Row labels (shown on the right) indicate the number of individuals in each cluster. Summary names for each of the clusters are provided on the left of the plot. To improve visualisation, only the most prevalent health conditions are shown. }\label{MMsummary1}
\end{figure}

\subsection{Full List of Health Conditions}

\begin{table}[!ht]
\caption{Table providing the full list of health conditions, and their coding used in the analysis and figures.} \label{LTCs}
    \centering
    \resizebox{0.87\textwidth}{!}{\begin{tabular}{|p{2.8cm}|p{4.8cm}|p{2.8cm}|p{4.8cm}|}
    \hline
    \rowcolor{LightGrey}
        Coding  & Condition & Coding  & Condition \\ \hline
        cancerall & Any cancer diagnosis & endometriosis & Endometriosis \\ \hline
        ca\_lung & Lung cancer & pcos & Polycystic ovary syndrome \\ \hline
        ca\_breast & Breast cancer & pernicious\_anaemia & Pernicious anaemia \\ \hline
        ca\_bowel & Bowel cancer & vte & Venous thromboembolism  \\ \hline
        ca\_prostate & Prostate cancer & pulmonary\_embolism & Pulmonary embolism \\ \hline
        lymphoma & Lymphoma & coagulopathy & Coagulopathy \\ \hline
        leukaemia & Leukaemia & depression & Depression \\ \hline
        ca\_skin & Skin cancer & anxiety & Anxiety \\ \hline
        melanoma & Melanoma & smi & Serious mental illness \\ \hline
        ca\_metastatic & Metastatic cancer & substance\_misuse & Substance misuse \\ \hline
        arrhythmia & Arrhythmia & alcohol\_problem & Alcohol problem \\ \hline
        af & Atrial fibrillation & adhd & Attention deficit hyperactivity disorder \\ \hline
        hypertension & Hypertension & eating\_disorder & Eating disorder \\ \hline
        hf & Hear failure & learning\_disability & Learning disability \\ \hline
        ihd & Ischemic heart disease & alzheimers & Alzheimer's disases \\ \hline
        valve\_disease & Heart valve disease & vascular\_dementia & Vascular dementia \\ \hline
        cardiomyopathy & Cardiomyopathy  & dementia & Dementia \\ \hline
        con\_heart\_disease & Congenital heart disease & parkinsons & Pakinson's disease \\ \hline
        pvd & Peripheral vascular disease & migraine & Migraine \\ \hline
        aortic\_aneurysm & Aortic aneurysm & ms & Multiple sclerosis \\ \hline
        tia & Transient ischaemic attack   & epilepsy & Epilepsy \\ \hline
        isch\_stroke & Ischaemic stroke & hemiplegia & Hemiplegia \\ \hline
        haem\_stroke & Haemorrhagic stroke & cfs & Chronic fatigue syndrome \\ \hline
        nos\_stroke & Stroke (not otherwise specified) & fibromyalgia & Fibromyalgia \\ \hline
        tia\_stroke & TIA stroke & oa & Osteoarthritis \\ \hline
        eczema & Eczema & osteoporosis & Osteoporosis \\ \hline
        psoriasis & Psoriasis & polymyalgia & Polymyalgia \\ \hline
        vitiligo & Vitiligo & ra & Rheumatoid arthritis \\ \hline
        alopecia & Alopecia & sjogren & Sjögren's syndrome  \\ \hline
        rhin\_conjunc & Rhinoconjunctivitis & sle & Systemic lupus erythematosus  \\ \hline
        sinusitis & Sinusitis & systematic\_sclerosis & Systematic sclerosis \\ \hline
        deaf & Deafness & psoriatic\_arthritis & Psoriatic arthritis \\ \hline
        blind & Blindness & ank\_spond & Ankylosing spondylitis \\ \hline
        cataract & Cataract & gout & Gout \\ \hline
        glaucoma & Glaucoma & ckd & Chronic kidney disease \\ \hline
        amd & Age-related macular degeneration & copd & Chronic obstructive pulmonary disease  \\ \hline
        retinopathy & Retinopathy & asthma & Asthma \\ \hline
        uveitis & Uveitis & osa & Obstructive sleep apnea \\ \hline
        scleritis & Scleritis & bronchiectasis & Bronchiectasis \\ \hline
        peptic\_ulcer & Peptic ulcer & pulmonary\_fibrosis & Pulmonary fibrosis \\ \hline
        ibd & Inflammatory bowel disease & hyperthyroidism & Hyperthyroidism \\ \hline
        ibs & Irritable bowel syndrme & hypothyroid & Hypothyroidism \\ \hline
        liver\_disease & Liver disease  & t1dm & Type 1 diabetes mellitus \\ \hline
        nash\_nafl & Nonalcoholic steatohepatitis/ Nonalcoholic fatty liver  & t2dm & Type 2 diabetes mellitus \\ \hline
        diverticuli & Diverticulitis & hiv & Human immunodeficiency virus \\ \hline
        coeliac & Coeliac disease & bph & Benign prostatic hyperplasia \\ \hline
        pancreatitis & Pancreatitis & erectile\_dysfunction & Erectile dysfunction \\ \hline
    \end{tabular}}
\end{table}

\end{document}